\documentclass[10pt,twocolumn,letterpaper]{article}

\usepackage{wacv}              

\definecolor{wacvblue}{rgb}{0.21,0.49,0.74}
\usepackage[pagebackref,breaklinks,colorlinks,allcolors=wacvblue]{hyperref}

\usepackage{tabularx}

\title{SNF-Bench: Separating Static Drift from Natural Flow in Long-Horizon Fixed-Camera Video Generation}

\author{
Matiur Rahman Minar\\
Sogang University\\
Seoul, Korea\\
{\tt\small minar@sogang.ac.kr}
\and
Seunghun Oh\\
Sogang University\\
Seoul, Korea\\
{\tt\small gnsgus190@sogang.ac.kr}
\and
Ganghyeon Jeong\\
Sogang University\\
Seoul, Korea\\
{\tt\small ganghyeon@sogang.ac.kr} 
\and
Unsang Park\\
Sogang University\\
Seoul, Korea\\
{\tt\small unsangpark@sogang.ac.kr}
}

\begin{document}
\maketitle

%
%
\IfFileExists{generated/macros.tex}{

\providecommand{\SNFdarNegN}{131}
\providecommand{\SNFdarNegPct}{8.3}
\providecommand{\SNFdarTotN}{1578}
\providecommand{\SNFddNBF}{+0.93}
\providecommand{\SNFfeatBackbone}{ORB}
\providecommand{\SNFflowBackbone}{RAFT}
\providecommand{\SNFnCategories}{6}
\providecommand{\SNFnIToVNative}{2}
\providecommand{\SNFnPromptsIToVSixty}{30}
\providecommand{\SNFnPromptsOneTwenty}{6}
\providecommand{\SNFnPromptsSixty}{23}
\providecommand{\SNFnPublicIToV}{7}
\providecommand{\SNFnPublicTTV}{7}
\providecommand{\SNFnPublicTTVPeers}{6}
\providecommand{\SNFrankCFDD}{1}
\providecommand{\SNFrankCFNBF}{7}
\providecommand{\SNFrankCFfBD}{7}
\providecommand{\SNFrankIFDD}{7}
\providecommand{\SNFrankIFNBF}{1}
\providecommand{\SNFrankIFfBD}{1}
\providecommand{\SNFspecVersions}{1.0, 1.1}
\providecommand{\SNFtopCatShare}{61}
\providecommand{\SNFvalCFDAR}{0.40}
\providecommand{\SNFvalCFDLR}{0.94}
\providecommand{\SNFvalCFNBF}{87.6}
\providecommand{\SNFvalCFfBD}{22.4}
\providecommand{\SNFvalIFDAR}{0.12}
\providecommand{\SNFvalIFDLR}{0.55}
\providecommand{\SNFvalIFNBF}{3.65}
\providecommand{\SNFvalIFfBD}{7.75}
\providecommand{\SNFvalTranslationDDRatio}{1.07}
\providecommand{\SNFvalTranslationFBDRatio}{1.86}
\providecommand{\SNFvalTranslationNBFRatio}{1.32}
}{%
  \GenericWarning{}{SNF-Bench: generated/macros.tex missing -- run
    scripts/paper_assets.py}}
\IfFileExists{generated/provenance.tex}{

\providecommand{\SNFprovenance}{%
Metrics computed under METRIC\_SPEC v1.0, 1.1 with the RAFT flow backbone and ORB features.}
}{%
  \GenericWarning{}{SNF-Bench: generated/provenance.tex missing -- run
    scripts/paper_assets.py}}

\begin{abstract}
Long-horizon video generation is evaluated with whole-frame metrics that reward motion and temporal consistency. For fixed-camera nature scenes this creates an ambiguity: motion of water, fire, smoke, or rain is desirable, whereas motion of the background is an error. A system can therefore score well on motion while its scene drifts, or on consistency while its flow stagnates. We introduce SNF-Bench, an evaluation framework for long-horizon fixed-camera generation that partitions each scene into static support and dynamic flow and reports static fidelity, flow persistence with absolute magnitude, and drift leakage separately, never as one score. Drift leakage is interpretive context rather than a headline measurement. Each factor is validated mechanistically rather than by correlation with preference: we inject global translation, rotation, and scale drift and progressive late freezing at known severity into real generations, and require each factor to respond in its stated direction and to remain selective against corruptions it does not target. Auditing publicly released long-horizon text-conditioned checkpoints under one recorded common inference configuration, plus an image-conditioned track with released-pipeline references and a deployment-sensitivity panel, we find that whole-frame motion and static-region drift induce near-opposite orderings of the same outputs. At maximum controlled translation, fBD and NBF rise to \SNFvalTranslationFBDRatio$\times$ and \SNFvalTranslationNBFRatio$\times$ baseline, but whole-frame Dynamic Degree reaches only \SNFvalTranslationDDRatio$\times$---rewarding the corruption. SNF-Bench measures where motion occurs and whether it persists; it does not measure physical realism. Project page: https://minar09.github.io/snfbench/
\end{abstract}

\section{Introduction}
\label{sec:intro}

Long-horizon, autoregressive (AR), and streaming video diffusion models now extend generation from seconds to minute-scale horizons~\cite{wang2025lingen,yesiltepe2025infinity,yang2025longlive,liu2025rolling,helios,yu2025videossm}. Evaluation has not kept pace: small temporal errors invisible in short clips accumulate over long rollouts into background displacement, color drift, or motion decay, and whole-frame metrics summarize these qualitatively different failures into one score.

\begin{figure*}[tb]
\centering
\includegraphics[width=\textwidth]{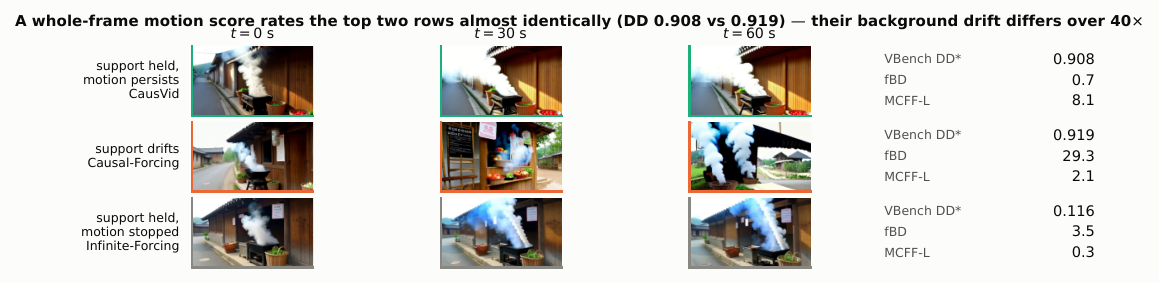}
\caption{\textbf{Why a fixed-camera benchmark needs a spatial decomposition.} Three $60$\,s outputs from released checkpoints under the recorded common configuration. A whole-frame motion score rates the first two rows almost identically, yet only the first holds its scene in place; their static-region drift differs by over an order of magnitude. The third row shows the opposite failure, which that score \emph{does} detect. DD${}^{*}$ is the per-video quantity underlying VBench~\cite{huang2023vbench} Dynamic Degree, computed under its published rule but read before the binarization the suite applies when aggregating, so that near-identical rows remain distinguishable as numbers.}
\label{fig:motivation}
\end{figure*}

The problem is sharpest in fixed-camera nature scenes. In a static-camera video of a river, the water should move while the bank, rocks, and surrounding landscape stay fixed. The same separation governs waterfalls, ocean waves, rainfall, fire, smoke, and wind-driven vegetation: some regions form \emph{static support} that must remain structurally fixed, while others carry \emph{dynamic flow} that must persist. Whole-frame motion measures reward optical flow regardless of its origin, crediting background drift as readily as genuine fluid motion, while temporal-consistency measures can favor videos whose stability comes from suppressing the flow they should sustain. Two opposite failures---static drift and flow decay---are thereby conflated, and increasingly so at long horizons, where accumulated background motion can dominate a sequence's apparent dynamics. Figure~\ref{fig:motivation} shows both failures in outputs of released checkpoints under the common setting.

We introduce \textbf{SNF-Bench} (Static--Natural Flow), an evaluation framework built around this decomposition, evaluated at $60$ and $120$\,s horizons with shorter ($5$\,s) and longer ($240$\,s) tiers retained as diagnostics. Measurement is organized into three factor groups, the third interpretive. \emph{Static fidelity} captures geometric displacement and residual flow inside nominally stationary regions. \emph{Flow persistence} reports the retention of dynamic-region motion with its absolute compensated magnitude, so a ratio cannot look strong merely because motion is uniformly weak. \emph{Drift leakage} relates static-region motion and global-motion compensation to dynamic-region flow; it contextualizes the two measurement axes rather than adding a third, and the central claim does not rest on it.

A benchmark is only as trustworthy as its measurements, so we validate the protocol independently of any model comparison. Controlled perturbations inject global translation, rotation, and scale drift, progressive motion attenuation, late freezing, photometric drift, and temporal repetition into real fixed-camera generations at known severity, and we require each factor to respond in the direction its definition states and to remain selective against corruptions it does not target. Because a parameter expressed in degrees is not commensurate with one expressed in pixels, geometric severity is specified as the mean displacement a corruption induces, which makes the families directly comparable. We additionally recompute every measurement under systematic erosion and dilation of the region partition.

We then audit publicly released long-horizon generators as executed through the recorded evaluation pipeline. The audit exposes cases where conventional metrics and SNF-Bench interpret the same outputs differently: high whole-frame motion coexisting with substantial static-region displacement, and high stability coexisting with decaying flow. We do not argue that existing metrics are incorrect. They answer a whole-frame question accurately; that question does not encode the spatial distinction this setting requires. Nor does SNF-Bench measure physical realism---a sequence may be drift-free and persistent yet physically implausible. The contribution is narrower and complementary: making the spatial origin and temporal persistence of motion explicit.

Our contributions are:
\begin{enumerate}
\item \textbf{SNF-Bench}, an evaluation framework for long-horizon fixed-camera generation organized around a static-support/dynamic-flow partition.
\item A \textbf{measurement protocol} that reports static fidelity and flow persistence with absolute magnitude as the primary axes, with drift leakage provided separately as interpretive context; the factors are never collapsed into one score.
\item \textbf{Mechanistic validation} of every reported factor under corruptions of known type and severity, with selectivity measured against the families each factor should ignore and every measurement recomputed under erosion and dilation of the region partition.
\item A \textbf{public-model audit} showing that this decomposition materially changes the interpretation and ordering of long-horizon behavior relative to whole-frame metrics.
\end{enumerate}

\section{Related Work}
\label{sec:related}

\paragraph{Video-generation evaluation.}
General-purpose suites assess visual quality, temporal consistency, motion, and semantic alignment. VBench~\cite{huang2023vbench} decomposes quality into dimensions including background consistency, motion smoothness and dynamic degree, and VBench-2.0~\cite{Zheng2025VBench20AV} extends it toward intrinsic faithfulness; EvalCrafter~\cite{eval_crafter} learns a weighting of many metrics from user opinion; Video-Bench~\cite{han2025videobench} scores generations with multimodal language models validated against human preference. Others specialize: T2V-CompBench~\cite{sun2024t2v_compbench} targets compositional binding, VMBench~\cite{ling2025vmbench} perceptually aligned motion quality, and WorldScore~\cite{duan2025worldscore} controllability and dynamics for world generation. These set the standards a benchmark paper must meet---explicit taxonomy, validated metrics, stated scale---but their measurements are whole-frame or sequence-level, and such an aggregate cannot reveal whether motion originates where intended or in displacement of nominally static structure. VMBench is the closest comparator, but it asks whether motion is perceptually plausible where we ask where it occurs and whether it survives compensation for global displacement. Our validation is correspondingly mechanistic---perturb by a known amount, require the expected response---complementing the human-alignment evidence those suites provide.

\paragraph{Long-horizon autoregressive generation.}
Autoregressive and streaming generators~\cite{chen2024diffusion,kim2024fifo,sun2025ardiffusion,zhou2025magi,lin2025stiv} extend synthesis beyond short bidirectional clips by producing frames or temporal chunks causally, including CausVid~\cite{yin2025causvid}, Self-Forcing~\cite{huang2025selfforcing}, Infinite-Forcing~\cite{infinite-forcing}, Rolling-Forcing~\cite{liu2025rolling}, Reward-Forcing~\cite{lu2025reward}, LongLive~\cite{yang2025longlive}, and Causal-Forcing~\cite{zhu2026causal} with its few-step successor Causal-Forcing++~\cite{zhao2026causal_pp}; bidirectional diffusion systems~\cite{blattmann2023svd,yang2025cogvideox,kong2024hunyuanvideo,hacohen2024ltxvideo,wan2025} provide short-horizon reference points. These systems differ substantially in temporal block size, training objective, memory and cache policy, and horizon-extension mechanism~\cite{yesiltepe2025infinity}, yet their evaluation typically relies on generic video metrics or method-specific qualitative analysis. That diversity motivates a recorded common-configuration audit, and the evaluation gap motivates a system-agnostic question: does static support remain fixed while dynamic flow persists?

\paragraph{Image dynamics and nature-flow animation.}
Work on animating still images has long modeled natural motion explicitly~\cite{holynski2021animating,xing2024dynamicrafter,ma2025cinemo,lei2025animateanything,li2025wonderplay}: controllable fluid animation separates static image structure from specified fluid motion~\cite{mahapatra2022controllable}, and Generative Image Dynamics models long-term scene motion through learned trajectory representations~\cite{Li_2024_gen_image_dynamics}. These works treat static support and dynamic content as distinct modeling targets; SNF-Bench elevates that distinction to an evaluation principle for long-horizon generators.

\section{SNF-Bench Design}
\label{sec:benchmark}

SNF-Bench evaluates long-horizon generation under a deliberately constrained setting: a fixed camera observing a scene containing both stationary structure and persistent natural motion. This isolates two failure modes that are difficult to separate in unconstrained video---\emph{static drift}, where nominally stationary structure moves over time, and \emph{flow decay}, where intended dynamics weaken or disappear. Every sequence is treated as a composition of two spatial roles, $\Omega_{\mathrm{static}}$ for regions expected to remain geometrically stationary and $\Omega_{\mathrm{flow}}$ for regions expected to exhibit sustained motion. The partition defines an evaluation role, not a semantic ontology: a riverbank is static support while the river surface is flow; buildings are static support in a rainfall scene while the falling rain is flow. SNF-Bench does not require dynamics inside $\Omega_{\mathrm{flow}}$ to follow any physical model. It measures whether motion stays localized to the appropriate regions and persists over time without mixing static support. Every system receives identical conditioning per item, and no prompt is rewritten for a particular method. Both conditioning modalities are evaluated under the same decomposition, differing only in how static support is established: T2V builds it in the opening sequence, while I2V takes it from a shared source image, so divergence there is attributable to the system rather than to a differently imagined scene. We report here both T2V and I2V at 60s and 120s; the 5s and 240s diagnostic tiers are provided in the supplementary.

\subsection{Scene Categories and Horizons}
\label{sec:categories}

Prompts are stratified across representative fixed-camera natural processes---channel water, ocean waves, precipitation, fire and smoke, lava, and wind-borne motion---with the category set fixed before the final sweep. The realized composition is dominated by directional and broad-area water motion with additional atmospheric and combustion categories, and conclusions should be read at that scope. The strata are not equally populated: channel water accounts for \SNFtopCatShare\% of the principal prompt set. Reported aggregates are means over prompts, not category macro-averages: three of the five populated categories hold two prompts each, so equal category weighting would buy balance at a cost in variance the sample cannot support. We instead recompute every system both ways and state which conclusions survive, with complete per-category counts in the supplementary.

Four temporal regimes are used: $5$\,s as an initialization check, $60$\,s as the primary reported horizon, $120$\,s as a supporting tier, and $240$\,s as a diagnostic extreme. Conclusions rest on the $60$\,s tier; the $120$\,s tier is reported alongside it as a supporting check, and the $5$ and $240$\,s tiers are in the supplementary. A run that cannot reach a target horizon under its declared track and setting is reported as unsupported rather than silently omitted.

\subsection{Region Partition}
\label{sec:masks}
The partition is obtained automatically and identically for every sequence. Mean optical-flow magnitude over the leading $12\%$ of sampled frame pairs is thresholded by Otsu's method~\cite{otsu1979threshold} into a dynamic region and its complement; the static region is then eroded and a $4\%$ frame border discarded, which together form an ignored transition band so that boundary pixels---where the partition is least certain and flow least reliable---contribute to neither region. Restricting the estimate to an early window reduces circularity from late accumulated drift: at $60$\,s the partition is fixed from the first $7.1$\,s and is therefore not defined by the late-window failure being measured. It still depends on early checkpoint behavior and on the same flow estimator used downstream. Two limitations bound the later conclusions. The partition is automatic, so a system drifting within the first $12\%$ of its rollout can influence its own partition; we therefore treat partitions as measured quantities and recompute every result under systematic erosion and dilation of the boundary (Sec.~\ref{sec:validation}). And a two-way partition cannot represent layered content: rain and snow transit static support, so those pixels are assigned by flow magnitude alone and static-fidelity values in precipitation scenes carry an unmodeled contribution. Derivation and sensitivity analyses are provided in the supplementary material; independent first-frame annotation remains an open extension.

\subsection{Benchmark Manifest}
\label{sec:manifest}
SNF-Bench is defined by a versioned manifest fixed before the final sweep, recording per item: identifier and category, exact prompt, target horizons, resolution and frame-rate metadata, and provenance. After freezing, items are never removed because a system performs poorly; corrections require a new benchmark version. Measurement coverage, partial records, and fBD abstentions are reported in Supp. Tables~S6--S7. Coverage is counted from valid per-video measurements rather than from the presence of a results file, so a run that terminated early is reported as partial.

\section{Drift--Motion Evaluation Protocol}
\label{sec:metrics}

SNF-Bench evaluates fixed-camera generation along three factor groups: \textbf{Static Fidelity} and \textbf{Flow Persistence} are the measurement axes, and \textbf{Drift Leakage} (DLR, DAR) contextualizes them rather than adding a third measurement. No individual score is sufficient: low background motion can be achieved by freezing the entire video, and high foreground motion can be produced by drifting the entire scene. We therefore always report static-region stability jointly with both the \emph{relative persistence} and \emph{absolute magnitude} of dynamic-region motion. Let $I_t$ denote frame $t$, $\Omega_{\mathrm{static}}$ and $\Omega_{\mathrm{flow}}$ the region masks of Sec.~\ref{sec:masks}, and $\mathbf{u}_t$ the optical flow from $I_t$ to $I_{t+1}$, estimated with RAFT~\cite{teed2020raft}. Each sequence defines an early window $\mathcal{E}$ and a late window $\mathcal{L}$ of equal duration, $\mathcal{E}$ beginning after a warm-up so that generation transients do not define the motion reference; extents are fixed in the specification.

\subsection{Static Fidelity}
\label{sec:static_metrics}

\paragraph{Feature-aligned Background Drift (fBD).}
ORB features~\cite{rublee2011orb} extracted inside $\Omega_{\mathrm{static}}$ from the first frame are matched against each late-window frame under robust geometric filtering; fBD~$(\downarrow)$ is the median matched-point displacement, averaged over the late window and expressed as a percentage of the frame diagonal. It captures accumulated geometric displacement of persistent structure and is intentionally insensitive to fine-scale flicker that moves no structure (See formal definition in Supp. S$1$).

\paragraph{Normalized Background Flow (NBF).}
Complementing accumulated drift, NBF~$(\downarrow)$ is the mean optical-flow magnitude inside the static mask, averaged over all frame pairs and normalized by frame width for resolution invariance and by the inter-frame interval $\Delta t$ for frame-rate invariance, giving units of $10^{-3}$ frame widths per second (Supp. S1). Here $\Delta t$ is the interval of the \emph{sampled} pairs actually used for flow estimation: methods are evaluated at their native frame rates, which differ across the roster, and the evaluation pipeline subsamples each clip to a common temporal rate before measurement. Expressing NBF per unit time rather than per frame makes the quantity independent of both the native rate and the sampling policy, so the measure remains comparable if either changes. NBF is a normalized magnitude, not a ratio: dividing by whole-frame flow would couple background leakage to foreground behavior and inflate the measure for methods whose dynamic motion decays.

\subsection{Flow Persistence}
\label{sec:flow_metrics}

\paragraph{Global-motion compensation.}
Coherent scene displacement is removed before dynamic-region motion is evaluated. A single translation vector is insufficient: under rotation or zoom the static-region flow field is not constant, and its component-wise median is near zero for a centered rotation, so a translation-only estimate would leave exactly the corruptions Sec.~\ref{sec:validation} injects uncompensated. We therefore estimate a robust global \emph{similarity} transform $T_t(\mathbf{x}) = s_t R_t \mathbf{x} + \mathbf{b}_t$ from static-region correspondences under RANSAC, and subtract the displacement field it induces from the measured flow pointwise, giving a compensated field $\tilde{\mathbf{u}}_t$ (Supp. S1).
A similarity rather than an unrestricted affine model is used because affine shear absorbs legitimate local deformation and over-corrects. Where correspondences are insufficient we fall back to median translation and record the fallback per sequence.

\paragraph{Motion-Compensated Foreground Flow (MCFF).}
MCFF~$(\uparrow)$ is the mean compensated flow magnitude inside $\Omega_{\mathrm{flow}}$ over a temporal window. We report it on both the early and the late window---MCFF-E and MCFF-L---so a method with uniformly negligible motion cannot appear strong through relative measures alone (Supp. S1).

\paragraph{Flow Persistence (FP).}
FP~$(\uparrow)$ is the ratio of late to early MCFF, clipped at a fixed constant to prevent unstable ratios when early motion is small (Supp. S1). FP near one indicates retained motion; substantially smaller values indicate decay; values above one are not automatically better. \textbf{FP is never reported without its MCFF magnitudes.}

\subsection{Drift Leakage and Attenuation}
\label{sec:drift_leakage}
Whole-frame metrics cannot determine whether apparent dynamic-region motion arises from intended flow or from displacement of the whole scene; we report two diagnostics, since neither subsumes the other. Let $F_{\mathrm{static}}$ denote mean flow magnitude over $\Omega_{\mathrm{static}}$, and $F_{\mathrm{raw}}$, $F_{\mathrm{comp}}$ the mean raw and compensated magnitude over $\Omega_{\mathrm{flow}}$, all in the late window, where drift has accumulated.

\paragraph{Drift Leakage Ratio (DLR).}
DLR~$(\downarrow)$, the ratio of static-region to raw dynamic-region flow magnitude in the late window, locates the sequence's motion energy (Supp. S1). It is not a fraction and is deliberately unbounded: $\mathrm{DLR}>1$ indicates that static support moves more than the region intended to move, a regime that occurs in practice.

\paragraph{Drift Attenuation Ratio (DAR).}
DAR~$(\downarrow)$, one minus the ratio of compensated to raw dynamic-region flow, measures the signed relative attenuation of measured dynamic-region flow after global similarity compensation (Supp. S1). Headline tables report DAR clipped to $[0,1]$ for compact comparison. Signed values are retained in the released records, and the incidence and magnitude of negative DAR are reported in Supp. Table~S14.

\paragraph{Interpretation.}
Neither is a causal decomposition of motion into drift and non-drift components. DLR and DAR are \emph{secondary diagnostics} that divide the work between them: DLR is a translational-leakage diagnostic whose rotational response is inverted, so rotational drift is carried by fBD and DAR instead, and that division is why DLR is never read alone. The underlying DAR is signed, and negative values are retained. High DLR with moderate DAR indicates static-region motion a \emph{rigid} model does not account for; it does not identify what the residual is.

\paragraph{Semantic guardrail.}
The decomposition is agnostic to \emph{what} moves: a sequence rendering drifting fog where a river was requested can occupy a favorable operating point. The natural guardrail is a prompt-alignment score---CLIP-based image--text similarity~\cite{hessel2021clipscore}, or a learned video-quality predictor~\cite{he2024videoscore} at higher cost. We do not report one, and the factors should be read accordingly: they establish where motion occurs and whether it persists, not that the depicted phenomenon is the one requested. Pairing SNF-Bench with a semantic score is how it should be used in practice.

\subsection{Joint Interpretation}
\label{sec:metric_interpretation}
No factor is read alone: low drift with low compensated motion indicates an over-stabilized sequence, high compensated motion with high static-region flow indicates motion contaminated by displacement, and high retention with negligible magnitude indicates persistent but trivial motion. SNF-Bench therefore reports no single scalar score, since collapsing the factors would recreate the ambiguity it exists to expose.

\paragraph{Implementation convention.}
Videos are normalized to a common evaluation resolution and subsampled to a common flow-estimation rate; no frame interpolation is performed, since synthesizing frames would alter the flow field being measured. Every estimator, threshold, window extent and constant is fixed globally before the sweep and applied unchanged to every system, and the per-sequence frame rate, resolution and compensation mode are released alongside the scores.

\section{Controlled Validation}
\label{sec:validation}

Before SNF-Bench is used to compare generators, we validate that its measurements respond to the failure modes they target. A benchmark is not justified merely because its metrics rank differently; each must change \emph{selectively}, \emph{monotonically}, and in the expected direction under known corruptions. We therefore construct controlled variants of fixed-camera reference clips with known corruption type and severity, independently of the audit in Sec.~\ref{sec:audit}: perturbation strengths, masks, windows, and hyperparameters are fixed before the final results are inspected.

\subsection{Perturbations and Criteria}
\label{sec:mechanistic_validation}

Given a reference sequence we generate variants that isolate individual components of the decomposition. \textbf{Geometric drift} applies a temporally increasing global transformation---translation, rotation, or scale---simulating camera-like background displacement that accumulates as long-horizon error does. \textbf{Motion attenuation} progressively blends dynamic-region content toward an early reference, and \textbf{late freezing} replaces all content after a chosen fraction of the sequence, both leaving static support untouched. \textbf{Photometric drift} varies brightness and color temperature while preserving geometry and motion, bounding the residual sensitivity of flow-based quantities. \textbf{Temporal repetition} replaces late dynamic content with a short repeated cycle. \textbf{Partition perturbation} erodes and dilates the region boundary rather than the video. Geometric severity is specified as the mean pixel displacement a corruption induces, with each family's parameter derived from it using the frame geometry. This is not presentational: a parameter in degrees is not commensurate with one in pixels, and at the evaluation resolution one degree of rotation induces only a few pixels of displacement, so families compared by their native parameters differ in injected severity by more than an order of magnitude. A factor is admitted only if its response direction matches its definition, that response is monotonic in severity, unrelated corruptions produce substantially smaller changes, and conclusions survive perturbation of the region partition. Factors failing these checks are reported as diagnostics rather than headline quantities. The severity specification and the admission outcome for every factor are in the supplementary.


\subsection{Validation Results}
\label{sec:validation_results}

\begin{figure*}[tb]
\centering
\includegraphics[width=\textwidth]{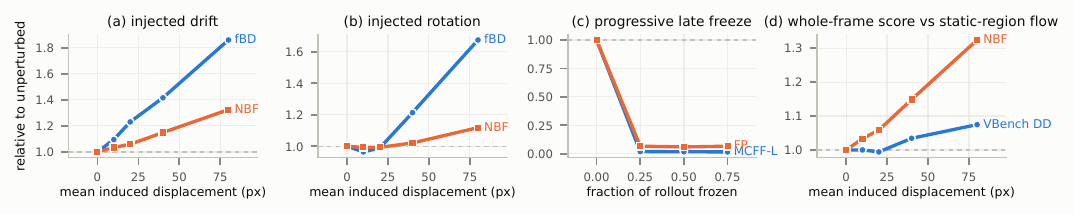}
\caption{\textbf{Mechanistic validation.} Curves are relative to unperturbed
fixed-camera clips. Translation raises fBD and NBF; progressive late freezing
collapses MCFF-L and FP; under the same translation, VBench Dynamic Degree
changes little and slightly upward. The perturbations are synthetic and
implicate no published method.}
\label{fig:validation}
\end{figure*}

Figure~\ref{fig:validation} and Table~\ref{tab:validation} report the outcome. Injected translation produces the expected monotone increase in both static-fidelity factors, and progressive late freezing drives MCFF-L and FP down---to $0.02$ and $0.07$ of their unperturbed values once the late window is frozen---while leaving the static factors comparatively unchanged, which is the selectivity the decomposition requires. At maximum translation, fBD and NBF reach \SNFvalTranslationFBDRatio$\times$ and \SNFvalTranslationNBFRatio$\times$ baseline, while VBench Dynamic Degree~\cite{huang2023vbench} reaches only \SNFvalTranslationDDRatio$\times$ and \emph{improves} on a corruption fixed-camera evaluation should penalize.

Two results qualify the framework rather than support it. Rotation is the harder geometric case for every factor, since correspondences degrade and the induced displacement is smallest at the frame center: both static-fidelity factors respond more weakly than under translation ($+0.70$ for fBD, $+0.60$ for NBF, against $+1.00$ for both), so neither is used alone. DLR is the sharper limitation---admitted on its target family, where it responds perfectly, but with an inverted rotational response ($-0.80$), so it is read only as a comparative diagnostic alongside DAR. Reported alongside these is the scope boundary of Sec.~\ref{sec:mechanistic_validation}: looping content sustains both persistence and magnitude, so SNF-Bench cannot distinguish genuine progression from repetition, and does not claim to. Table~\ref{tab:robustness_main} summarizes the complementary partition checks;
the full per-checkpoint sensitivity analysis is reported in the supplementary.

\begin{table}[tb]
\centering
\scriptsize
\setlength{\tabcolsep}{1.5pt}
\begin{tabular}{llcccc}
\toprule
Factor & target & $\rho$ & $\rho_{\mathrm{rot}}$ & off-target & admitted \\
\midrule
fBD & drift & +1.00 & +0.70 & 14\% & \checkmark \\
NBF & drift & +1.00 & +0.60 & 6\% & \checkmark \\
MCFF-L & freeze & +1.00 & n/a & 14\% & \checkmark \\
FP & freeze & +0.80 & n/a & 21\% & \checkmark \\
DLR & translational leakage & +1.00 & -0.80 & 5\% & \checkmark \\
DAR & drift & +0.90 & +1.00 & 18\% & \checkmark \\
VBench DD & drift & +0.70 & +0.70 & 2\% & -- \\
\bottomrule
\end{tabular}
\caption{\textbf{Mechanistic validation and factor admission.} $\rho$ is signed target-family Spearman correlation; $\rho_{\mathrm{rot}}$ reports rotational response; \emph{off-target} is the largest relative excursion under photometric and mask perturbations. Admission requires $\rho\geq0.7$ and off-target response below $25\%$. Rotation is reported but not gated: DLR is inverted there, while fBD and DAR carry that case. VBench DD is a contrast row, not a candidate factor. Twelve clips per severity.}
\label{tab:validation}
\end{table}
\begin{table}[tb]
\centering
\scriptsize
\setlength{\tabcolsep}{4pt}
\begin{tabular}{p{0.51\linewidth} p{0.40\linewidth}}
\toprule
Check & result \\
\midrule
Static-mask area & 0.614--0.663 \\
Area--fBD Spearman & $\rho=-0.75$, exact $p=0.066$, $n=7$ \\
No precipitation: NBF order & identical \\
No precipitation: fBD extremes & unchanged \\
Boundary perturbation: max. response & $\leq 14\%$ \\
\bottomrule
\end{tabular}
\caption{\textbf{Partition robustness at 60\,s.} Static-region area is similar across checkpoints, and the headline NBF ordering is unchanged when precipitation is excluded. The negative area--fBD association has the direction expected under early-mask circularity and is therefore reported as an open validity concern rather than dismissed.}
\label{tab:robustness_main}
\end{table}

\section{Public-Model Audit}
\label{sec:audit}

We apply SNF-Bench to publicly released long-horizon generators. The goal is not a universal leaderboard but a determination of how checkpoints occupy the static-fidelity/flow-persistence space under one recorded common T2V configuration.

\paragraph{Systems and setting.}
The audit covers \SNFnPublicTTV{} publicly released autoregressive or streaming generators with accessible checkpoints, frozen as a set before the sweep. Every T2V checkpoint is loaded into the same recorded inference implementation rather than its released pipeline. Fixed before the sweep to give the distilled checkpoint family a shared step budget, the configuration uses a four-step denoising schedule (indices warped by the released scheduler), six frames per block, seed $0$, and one rollout/cache path. The T2V runs also share $832\!\times\!480$ output at $16$ fps. Within a horizon, checkpoint weights and their recorded regular/EMA selection are the only system-specific inference inputs. Accordingly, these results characterize checkpoints under this common configuration, not each method's published inference procedure; a checkpoint distilled for another step budget is not observed at its own operating point. The I2V track instead separates released-pipeline outputs from dagger-marked long-horizon-wrapper rows (Sec.~\ref{sec:results_i2v}).

\paragraph{Controls and statistics.}
Within each item every system receives identical conditioning, and the seed is fixed before the sweep. This makes each run reproducible but does not make different checkpoints' stochastic trajectories equivalent; cross-system comparison rests on the paired-item design. Videos are retained in their recorded form and the normalization of Sec.~\ref{sec:metrics} applied unchanged. Scores are computed per item and averaged over items; the sensitivity of every conclusion to category macro-averaging is reported in the supplementary. Uncertainty is estimated by percentile bootstrap~\cite{efron1979bootstrap} over items, and we report $95\%$ marginal intervals. These intervals summarize per-checkpoint uncertainty; all headline separation claims use paired bootstrap distributions of per-prompt differences, reported in Supp. Table~S3. Measurement coverage and fBD abstentions are reported in Supp. Tables~S6--S7; incomplete metric records are counted as partial rather than silently treated as successful. To test whether the decomposition changes interpretation, the same outputs are additionally scored with established whole-frame metrics and the induced orderings compared. The focus is interpretation change, not a claim that existing metrics are wrong.

\section{Results and Interpretation Changes}
\label{sec:results}

Unless stated otherwise, T2V comparisons refer to checkpoints under the recorded common configuration. I2V released-pipeline and wrapper results are identified separately and never ranked across settings.

\subsection{Recorded Common-Configuration T2V Audit}
\label{sec:results_t2v}

Table~\ref{tab:t2v_audit} reports the primary $60$\,s T2V tier and a $120$\,s directional check. Because the $120$\,s panel has \SNFnPromptsOneTwenty{} different prompts, it is not compared column-by-column with $60$\,s. The $5$ and $240$\,s diagnostics and all four horizons are in the supplementary. Figure~\ref{fig:t2v_qualitative} shows every audited checkpoint on one prompt at t={0,30,60} s, ordered by fBD; the measured differences in static-region drift are visually apparent.

\begin{figure*}[t]
\centering
\includegraphics[width=\textwidth]{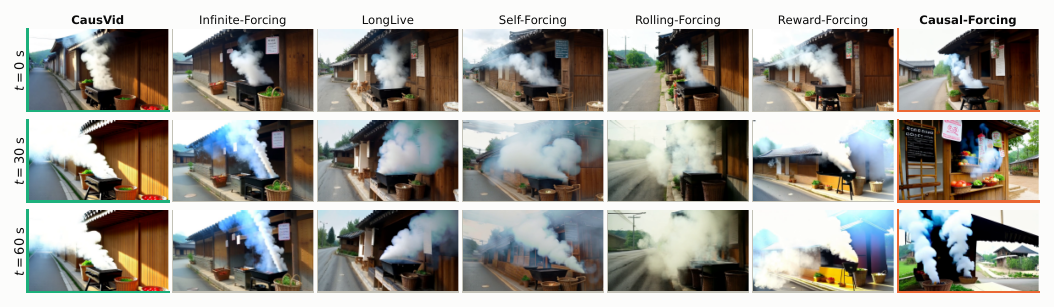}
\caption{\textbf{T2V comparison under the recorded common configuration.} Rows show t={0,30,60} s for one shared prompt; columns are ordered by increasing measured fBD. Green and orange borders identify the CausVid/Causal-Forcing pair highlighted in Figure 1; the fixed selection rule is documented in Supplementary Section S14.}
\label{fig:t2v_qualitative}
\end{figure*}

\begin{table}[!tb]
\centering
\scriptsize
\setlength{\tabcolsep}{2.5pt}
\begin{tabularx}{\linewidth}{X c c c c}
\toprule
Method & fBD$\downarrow$ & NBF$\downarrow$ & MCFF E$\rightarrow$L & FP$\uparrow$ \\
\midrule
\multicolumn{5}{l}{\textit{60s horizon  ($n=23$)}} \\
CausVid & 8.08 $\pm$ 3.15 & 9.27 $\pm$ 4.14 & 2.87\,$\rightarrow$\,1.96 & 0.67 $\pm$ 0.18 \\
Self-Forcing & 11.4 $\pm$ 3.63 & 12.7 $\pm$ 3.97 & 10.4\,$\rightarrow$\,2.54 & 0.44 $\pm$ 0.22 \\
Infinite-Forcing & 7.75 $\pm$ 3.84 & 3.65 $\pm$ 0.88 & 1.17\,$\rightarrow$\,0.29 & 0.63 $\pm$ 0.22 \\
Rolling-Forcing & 16.8 $\pm$ 3.69 & 11.2 $\pm$ 2.70 & 3.18\,$\rightarrow$\,1.57 & 0.75 $\pm$ 0.29 \\
Reward-Forcing & 14.8 $\pm$ 3.71 & 8.51 $\pm$ 2.79 & 3.81\,$\rightarrow$\,0.80 & 0.49 $\pm$ 0.18 \\
LongLive & 15.9 $\pm$ 2.89 & 13.9 $\pm$ 4.76 & 6.46\,$\rightarrow$\,2.41 & 0.75 $\pm$ 0.28 \\
Causal-Forcing & 22.4 $\pm$ 2.26 & 87.6 $\pm$ 18.6 & 9.84\,$\rightarrow$\,7.21 & 0.83 $\pm$ 0.31 \\
\midrule
\multicolumn{5}{l}{\textit{120s horizon  ($n=6$)}} \\
CausVid & 6.15 & 3.42 & 0.36\,$\rightarrow$\,0.52 & 1.06 \\
Self-Forcing & 21.2 & 6.80 & 4.85\,$\rightarrow$\,0.82 & 0.85 \\
Infinite-Forcing & 16.1 & 2.60 & 0.50\,$\rightarrow$\,0.32 & 0.64 \\
Rolling-Forcing & 23.8 & 10.6 & 3.33\,$\rightarrow$\,0.77 & 0.38 \\
Reward-Forcing & 20.0 & 11.6 & 2.05\,$\rightarrow$\,0.69 & 0.39 \\
LongLive & 22.5 & 9.66 & 5.10\,$\rightarrow$\,1.19 & 0.81 \\
Causal-Forcing & 21.8 & 130 & 10.5\,$\rightarrow$\,10.7 & 0.90 \\
\bottomrule
\end{tabularx}
\caption{\textbf{Recorded common-configuration T2V audit.} A four-step denoising schedule with released-scheduler warping, six frames/block, and seed $0$. Header $n$ is the maximum valid item count within each horizon; factor-specific abstentions are in Table~S6. Units: fBD is \% frame diagonal; NBF is $10^{-3}$ frame widths/s; MCFF is px/sampled interval; FP is unitless. FP is the mean of per-video clipped late/early ratios and therefore need not equal the ratio of the displayed aggregate MCFF-L and MCFF-E means. Means with percentile-bootstrap 95\% CI half-width shown for compactness (10k resamples, seed 0); $n<10$ shows means only. MCFF-E$\rightarrow$L accompanies FP. Separation uses paired per-prompt intervals (Table~S3). DLR/DAR are supplementary.}
\label{tab:t2v_audit}
\end{table}
\begin{table}[!tb]
\centering
\scriptsize
\setlength{\tabcolsep}{3pt}
\begin{tabularx}{\linewidth}{X c c c c}
\toprule
Method & fBD$\downarrow$ & NBF$\downarrow$ & MCFF E$\rightarrow$L & FP$\uparrow$ \\
\midrule
\multicolumn{5}{l}{\textit{60s horizon  ($n=30$)}} \\
Causal-Forcing++ (2-step)$^{\dagger}$ & 9.18 & 5.97 & 0.85\,$\rightarrow$\,0.60 & 0.86 \\
Causal-Forcing++ (1-step)$^{\dagger}$ & 5.34 & 7.05 & 0.86\,$\rightarrow$\,0.57 & 0.77 \\
Causal-Forcing (frame-wise)$^{\dagger}$ & 21.1 & 36.4 & 2.80\,$\rightarrow$\,2.69 & 0.89 \\
Causal-Forcing++ (2-step, frame-wise) & 9.94 & 6.91 & 1.30\,$\rightarrow$\,0.78 & 1.02 \\
Self-Forcing$^{\dagger}$ & 17.3 & 5.52 & 1.50\,$\rightarrow$\,0.69 & 0.89 \\
CausVid$^{\dagger}$ & 19.9 & 3.31 & 0.58\,$\rightarrow$\,0.41 & 0.81 \\
Causal-Forcing++ (1-step, frame-wise) & 10.4 & 5.25 & 0.53\,$\rightarrow$\,0.52 & 1.01 \\
\midrule
\multicolumn{5}{l}{\textit{120s horizon  ($n=20$)}} \\
Causal-Forcing++ (2-step)$^{\dagger}$ & 9.65 & 8.81 & 0.89\,$\rightarrow$\,0.82 & 0.93 \\
Causal-Forcing++ (1-step)$^{\dagger}$ & 7.26 & 9.39 & 0.84\,$\rightarrow$\,0.66 & 0.86 \\
Causal-Forcing (frame-wise)$^{\dagger}$ & 19.9 & 52.7 & 4.82\,$\rightarrow$\,5.06 & 1.08 \\
Causal-Forcing++ (2-step, frame-wise) & 11.4 & 17.3 & 1.89\,$\rightarrow$\,1.54 & 1.01 \\
Self-Forcing$^{\dagger}$ & 23.0 & 10.3 & 2.41\,$\rightarrow$\,0.76 & 0.45 \\
CausVid$^{\dagger}$ & 19.7 & 6.04 & 1.15\,$\rightarrow$\,0.83 & 0.67 \\
Causal-Forcing++ (1-step, frame-wise) & 13.6 & 7.39 & 1.01\,$\rightarrow$\,0.35 & 0.64 \\
\bottomrule
\end{tabularx}
\caption{\textbf{I2V released pipelines and deployment sensitivity at 60 and 120\,s.} Unmarked rows are released-pipeline outputs; $^{\dagger}$ rows use the wrapper; settings are not compared. Header $n$ is the maximum valid item count within each horizon; factor-specific abstentions are in Table~S6. Units: fBD is \% frame diagonal; NBF is $10^{-3}$ frame widths/s; MCFF is px/sampled interval; FP is dimensionless. FP is the mean of per-video clipped late/early ratios and therefore need not equal the ratio of the displayed aggregate MCFF-L and MCFF-E means. Means are shown; DLR and DAR are reported in Supplementary Table S9. MCFF-E→L accompanies FP, and released-pipeline and wrapper rows are not compared.}
\label{tab:i2v_audit}
\end{table}

\paragraph{Motion persistence.} We read $\bigl(\mathrm{MCFF}(\mathcal{L}),\,\mathrm{FP}\bigr)$ jointly: sustained flow needs non-trivial late motion and high retention. MCFF-E beside MCFF-L exposes uniformly negligible motion.

\paragraph{Operating regimes.} Joint static-fidelity and persistence columns expose low-drift/low-motion, high-motion/high-drift, and low-drift/non-trivial-motion regimes without a scalar score. Infinite-Forcing~\cite{infinite-forcing} has the lowest NBF under the recorded common configuration---paired per-prompt intervals versus all \SNFnPublicTTVPeers{} peers exclude zero (Supp. Table~S3)---but also the least surviving late motion. Causal-Forcing~\cite{zhu2026causal} occupies the opposite, higher-background-flow corner.

\subsection{Whole-Frame Motion and Static-Region Flow Yield Different Interpretations}
\label{sec:results_drift}

Across \SNFnPublicTTV{} checkpoint means, Dynamic Degree tracks static-region flow at $\rho=\SNFddNBF$ (descriptive; per-item analysis in the supplementary): more apparent motion also tends to mean more background motion. Causal-Forcing ranks \SNFrankCFDD{}/\SNFnPublicTTV{} on Dynamic Degree but \SNFrankCFNBF{}/\SNFnPublicTTV{} on NBF; Infinite-Forcing shows the reverse, ranking \SNFrankIFDD{}/\SNFnPublicTTV{} on Dynamic Degree and \SNFrankIFNBF{}/\SNFnPublicTTV{} on NBF. Causal-Forcing's DLR of \SNFvalCFDLR{} indicates static-region motion energy nearly equal to dynamic-region energy. Its DAR of \SNFvalCFDAR{} means fitted similarity compensation reduces measured dynamic-region flow by that proportion; it is attenuation, not causal decomposition. The residual is unidentified and may include non-rigid instability, intended motion, partition contamination, or flow error.

\subsection{Image-Conditioned (I2V) Track: Released Pipelines and Deployment Sensitivity}
\label{sec:results_i2v}

\begin{figure}[t]
\centering
\includegraphics[width=\linewidth]{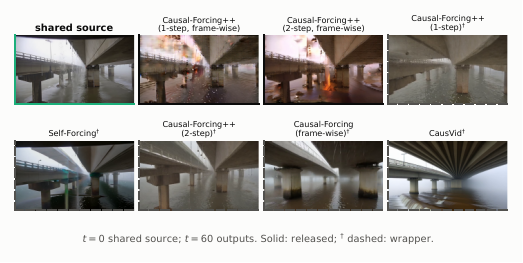}
\caption{\textbf{I2V comparison (60s) on a shared source (0s).} Unmarked outputs are produced by released pipelines; † marks outputs generated with the common wrapper.}
\label{fig:i2v_qualitative}
\end{figure}

At $60$ and $120$\,s, Table~\ref{tab:i2v_audit} reports two released-pipeline rows and five dagger-marked wrapper rows per horizon; the two settings are kept separate and are never ranked together. The panels show deployment sensitivity without attributing its cause. Figure~\ref{fig:i2v_qualitative} compares a distinct $60$\,s sample: each system shares one source image, so static support is given rather than imagined differently. Complete horizons and $5$\,s bidirectional references are in the supplementary.

\section{Limitations and Conclusion}
\label{sec:limitations}

SNF-Bench is limited to long-horizon fixed-camera generation where static support and dynamic flow separate. It measures neither physical nor semantic realism, cannot distinguish progression from repetition, and depends on an automatic partition bounded by our erosion--dilation study. 

Whole-frame motion and region-resolved static fidelity can materially change the interpretation of the same outputs. Reporting static fidelity and flow persistence separately, with drift quantities only as context, complements broad benchmarks without a composite score. The released specification and common-configuration checkpoint audit provide a reproducible basis for evaluation; independent partitions and broader non-water coverage remain important next steps.

{
    \small
    \bibliographystyle{ieeenat_fullname}
    \bibliography{main}
}

%
%
%
\clearpage
\maketitlesupplementary

\appendix
\renewcommand{\thesection}{S\arabic{section}}
\renewcommand{\thefigure}{S\arabic{figure}}
\renewcommand{\thetable}{S\arabic{table}}
\renewcommand{\theequation}{S\arabic{equation}}

\setcounter{section}{0}
\setcounter{figure}{0}
\setcounter{table}{0}
\setcounter{equation}{0}

% ...

\renewcommand{\thesection}{S\arabic{section}}
\renewcommand{\thefigure}{S\arabic{figure}}
\renewcommand{\thetable}{S\arabic{table}}
\renewcommand{\theequation}{S\arabic{equation}}\

\IfFileExists{generated/macros.tex}{

\providecommand{\SNFdarNegN}{131}
\providecommand{\SNFdarNegPct}{8.3}
\providecommand{\SNFdarTotN}{1578}
\providecommand{\SNFddNBF}{+0.93}
\providecommand{\SNFfeatBackbone}{ORB}
\providecommand{\SNFflowBackbone}{RAFT}
\providecommand{\SNFnCategories}{6}
\providecommand{\SNFnIToVNative}{2}
\providecommand{\SNFnPromptsIToVSixty}{30}
\providecommand{\SNFnPromptsOneTwenty}{6}
\providecommand{\SNFnPromptsSixty}{23}
\providecommand{\SNFnPublicIToV}{7}
\providecommand{\SNFnPublicTTV}{7}
\providecommand{\SNFnPublicTTVPeers}{6}
\providecommand{\SNFrankCFDD}{1}
\providecommand{\SNFrankCFNBF}{7}
\providecommand{\SNFrankCFfBD}{7}
\providecommand{\SNFrankIFDD}{7}
\providecommand{\SNFrankIFNBF}{1}
\providecommand{\SNFrankIFfBD}{1}
\providecommand{\SNFspecVersions}{1.0, 1.1}
\providecommand{\SNFtopCatShare}{61}
\providecommand{\SNFvalCFDAR}{0.40}
\providecommand{\SNFvalCFDLR}{0.94}
\providecommand{\SNFvalCFNBF}{87.6}
\providecommand{\SNFvalCFfBD}{22.4}
\providecommand{\SNFvalIFDAR}{0.12}
\providecommand{\SNFvalIFDLR}{0.55}
\providecommand{\SNFvalIFNBF}{3.65}
\providecommand{\SNFvalIFfBD}{7.75}
\providecommand{\SNFvalTranslationDDRatio}{1.07}
\providecommand{\SNFvalTranslationFBDRatio}{1.86}
\providecommand{\SNFvalTranslationNBFRatio}{1.32}
}{%
  \GenericWarning{}{SNF-Bench: generated/macros.tex missing}}
\IfFileExists{generated/provenance.tex}{

\providecommand{\SNFprovenance}{%
Metrics computed under METRIC\_SPEC v1.0, 1.1 with the RAFT flow backbone and ORB features.}
}{%
  \GenericWarning{}{SNF-Bench: generated/provenance.tex missing}}

\noindent
This supplement records the material supporting the main paper without being
required to follow its argument: the region-partition procedure and its
sensitivity study, complete audit tables for both tracks at every horizon,
measurement coverage, protocols for the two complementary perceptual checks, and
additional qualitative material. The text-conditioned and image-conditioned
tracks are reported here at equal depth. Section numbering is independent of the
main paper. We release the benchmark code and data here: https://minar09.github.io/snfbench/

\section{Formal Factor Definitions}
\label{sup:defs}

The definitions below are those referenced from the main paper's protocol
section. Notation follows it: $I_t$ is frame $t$ of a $T$-frame sequence,
$\Omega_{\mathrm{static}}$ and $\Omega_{\mathrm{flow}}$ are the region masks,
$\mathbf{u}_t$ is the optical flow from $I_t$ to $I_{t+1}$, and $\mathcal{E}$
and $\mathcal{L}$ are the early and late windows.

\paragraph{Feature-aligned Background Drift.}
For matched ORB pairs $\mathcal{M}_t$ between the first frame and frame $t$,
\begin{align}
D_{\mathrm{feat}}(t) &= \operatorname*{median}_{(p,q)\in\mathcal{M}_t} \lVert p-q \rVert_2, \nonumber\\
\mathrm{fBD} &= \frac{100}{d}\cdot\frac{1}{|\mathcal{L}|}\sum_{t\in\mathcal{L}} D_{\mathrm{feat}}(t) \;\;(\downarrow),
\end{align}
with $d$ the frame diagonal, so fBD is a percentage of image size.

\paragraph{Normalized Background Flow.}
\begin{equation}
\mathrm{NBF} = \frac{10^3}{W\,\Delta t}\cdot\frac{1}{T-1}\sum_{t=1}^{T-1}
\frac{1}{|\Omega_{\mathrm{static}}|}\sum_{\mathbf{x}\in\Omega_{\mathrm{static}}}
\lVert \mathbf{u}_t(\mathbf{x}) \rVert_2
\;\;(\downarrow),
\end{equation}
with $W$ the frame width and $\Delta t$ the interval of the \emph{sampled} pairs
actually used for flow estimation, so the quantity is independent of both the
native frame rate and the sampling policy.

\paragraph{Global-motion compensation.}
With $T_t(\mathbf{x}) = s_t R_t \mathbf{x} + \mathbf{b}_t$ the robust similarity
transform fitted to static-region correspondences and
$\mathbf{d}_t(\mathbf{x}) = T_t(\mathbf{x})-\mathbf{x}$ the displacement field it
induces,
\begin{equation}
\tilde{\mathbf{u}}_t(\mathbf{x}) = \mathbf{u}_t(\mathbf{x}) - \mathbf{d}_t(\mathbf{x}).
\end{equation}

\paragraph{Motion-Compensated Foreground Flow and Flow Persistence.}
\begin{equation}
\mathrm{MCFF}(\mathcal{W}) = \frac{1}{|\mathcal{W}|}\sum_{t\in\mathcal{W}}
\frac{1}{|\Omega_{\mathrm{flow}}|}\sum_{\mathbf{x}\in\Omega_{\mathrm{flow}}}
\lVert \tilde{\mathbf{u}}_t(\mathbf{x}) \rVert_2
\;\;(\uparrow),
\end{equation}
\begin{equation}
\mathrm{FP} = \min\!\left(\frac{\mathrm{MCFF}(\mathcal{L})}{\mathrm{MCFF}(\mathcal{E})+\epsilon},\, c\right)
\;\;(\uparrow),
\end{equation}
clipped at a fixed $c$ so that a vanishing early window cannot produce an
unbounded ratio.

\paragraph{Drift leakage and attenuation.}
With $F_{\mathrm{static}}$ the mean flow magnitude over $\Omega_{\mathrm{static}}$
and $F_{\mathrm{raw}}$, $F_{\mathrm{comp}}$ the mean raw and compensated
magnitudes over $\Omega_{\mathrm{flow}}$, all in the late window,
\begin{equation}
\mathrm{DLR} = \frac{F_{\mathrm{static}}}{F_{\mathrm{raw}}+\epsilon}\;\;(\downarrow),
\qquad
\mathrm{DAR} = 1 - \frac{F_{\mathrm{comp}}}{F_{\mathrm{raw}}+\epsilon}\;\;(\downarrow).
\end{equation}
DLR is not a fraction and is deliberately unbounded: $\mathrm{DLR}>1$ indicates
static support moving more than the region intended to move. DAR is stored
signed and reported clipped to $[0,1]$.

\section{Region Partition}
\label{sup:masks}

\paragraph{Derivation.}
For every generated sequence we average optical-flow magnitude~\cite{teed2020raft}
over the leading $12\%$ of sampled frame pairs and threshold it by Otsu's
method~\cite{otsu1979threshold} into a dynamic region and its complement. The
static region is then reduced by a fixed morphological erosion and a $4\%$ frame
border discarded; together these form the ignored transition band, so boundary
pixels---where the partition is least certain and flow least reliable---contribute
to neither region. The same procedure applies to both tracks.

\paragraph{Why the estimate is restricted to an early window.}
Deriving a partition from a sequence's entire motion field would be circular,
since accumulated background drift would itself be read as evidence that the
affected region is dynamic. Restricting the estimate to the early window reduces
circularity from late accumulated drift: at the $60$\,s horizon the partition
is fixed from the first $7.1$\,s and is not defined by the late-window failure
being measured. It nevertheless remains dependent on early checkpoint behavior
and on the same flow estimator used downstream.

\paragraph{Limitations.}
The partition is obtained automatically rather than by annotation, so no
inter-annotator agreement is reported, and for the image-conditioned track it is
derived per generated sequence rather than once from the shared source image.
The early window is a weaker guarantee than a first-frame annotation: a system
that drifts within the first $12\%$ of its rollout can influence its own
partition. We therefore treat partitions as measured quantities and report the
consequence directly---every measurement is recomputed under systematic erosion
and dilation of the boundary, with the response in Table~\ref{tab:validation}.

\begin{figure*}[tb]
\centering
\includegraphics[width=\linewidth]{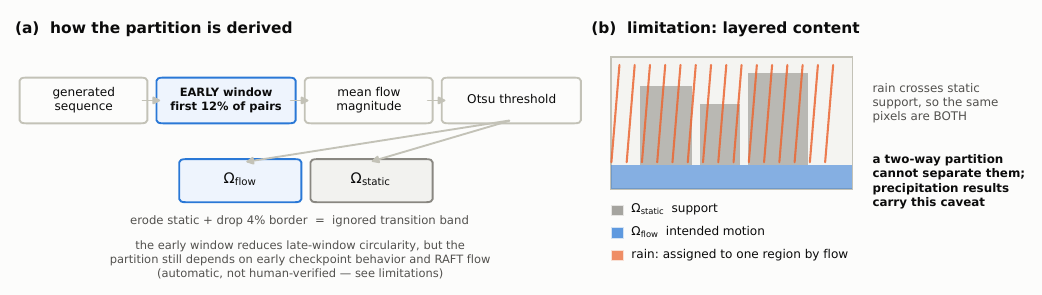}
\caption{\textbf{Region-partition protocol.} (a)~The partition is derived automatically from early-window flow, with an ignored transition band at the boundary. (b)~A two-way partition cannot represent layered content such as rain crossing static support, which is the scope limit stated in the main paper.}
\label{sup:fig_mask}
\end{figure*}

\section{Validation Protocol}
\label{sup:validation}

Geometric severity is specified in a common physical unit---the mean pixel
displacement a corruption induces over the frame by the end of the
rollout---and each family's parameter is derived from it using the frame
geometry: a translation of $d$ pixels, a rotation of $d/\bar{r}$ radians, or a
zoom of $1 + d/\bar{r}$, where $\bar{r}$ is the mean distance from the frame
center. This is what makes the geometric families comparable. A parameter in
degrees is not commensurate with one in pixels: at the evaluation resolution one
degree of rotation induces roughly $2.7$\,px of mean displacement, so families
compared by their native parameters differ in injected severity by more than an
order of magnitude, and a factor can appear non-monotone simply because the
corruption was too weak to clear the clip's own baseline drift.

\begin{table}[tb]
\centering
\scriptsize
\setlength{\tabcolsep}{1.5pt}
\begin{tabular}{llcccc}
\toprule
Factor & target & $\rho$ & $\rho_{\mathrm{rot}}$ & off-target & admitted \\
\midrule
fBD & drift & +1.00 & +0.70 & 14\% & \checkmark \\
NBF & drift & +1.00 & +0.60 & 6\% & \checkmark \\
MCFF-L & freeze & +1.00 & n/a & 14\% & \checkmark \\
FP & freeze & +0.80 & n/a & 21\% & \checkmark \\
DLR & translational leakage & +1.00 & -0.80 & 5\% & \checkmark \\
DAR & drift & +0.90 & +1.00 & 18\% & \checkmark \\
VBench DD & drift & +0.70 & +0.70 & 2\% & -- \\
\bottomrule
\end{tabular}
\caption{\textbf{Mechanistic validation and factor admission.} $\rho$ is signed target-family Spearman correlation; $\rho_{\mathrm{rot}}$ reports rotational response; \emph{off-target} is the largest relative excursion under photometric and mask perturbations. Admission requires $\rho\geq0.7$ and off-target response below $25\%$. Rotation is reported but not gated: DLR is inverted there, while fBD and DAR carry that case. VBench DD is a contrast row, not a candidate factor. Twelve clips per severity.}
\label{tab:validation}
\end{table}

\section{Partition and Composition Robustness}
\label{sup:robustness}

Two questions decide whether the region partition can be trusted as a
measurement basis: whether a system can be advantaged by receiving an easier
mask, and whether the ordering depends on the one scene category the partition
cannot represent.

The static region occupies a similar share of the frame for every audited
system, between $0.61$ and $0.66$ on average, so no system is scored over a
markedly larger or smaller support than another. Across systems, that area
correlates negatively with feature-based drift. Over seven systems the
correlation is not determined, but its sign is the one circularity would
produce---a system that drifts within the estimation window enlarges its own
dynamic region and shrinks the support its drift is measured over---and we
report it as an open validity question rather than dismissing it on
significance. The decisive test, recomputing every score under a region
partition not derived from the same sequence, is stated as required future work
in Sec.~\ref{sup:masks}; we do not claim to have run it.

Removing precipitation, where rain and snow cross static support and a two-way
partition must assign those pixels to one region by flow magnitude alone, leaves
the normalized-background-flow ordering unchanged and moves only mid-field
positions under feature-based drift.

\begin{table*}[tb]
\centering
\small
\begin{tabular}{l c}
\toprule
System & static-region area \\
\midrule
CausVid & 0.663 $\pm$ 0.073 \\
Reward-Forcing & 0.661 $\pm$ 0.060 \\
Self-Forcing & 0.651 $\pm$ 0.087 \\
Infinite-Forcing & 0.650 $\pm$ 0.090 \\
LongLive & 0.644 $\pm$ 0.095 \\
Rolling-Forcing & 0.616 $\pm$ 0.107 \\
Causal-Forcing & 0.614 $\pm$ 0.086 \\
\midrule
\multicolumn{2}{l}{\emph{Does mask area predict the drift score?}} \\
Spearman(area, fBD) over 7 systems & -0.75, exact $p=0.066$ \\
\midrule
\multicolumn{2}{l}{\emph{Ordering with precipitation excluded (3 of 23 prompts)}} \\
fBD ordering & best and worst unchanged \\
NBF ordering & identical \\
\bottomrule
\end{tabular}
\caption{\textbf{Partition and composition robustness.} The static region occupies a similar share of the frame for every audited system, so no system is scored over a much larger or smaller support than another. The area does correlate negatively with drift across systems; over seven systems that correlation is not determined, but its sign is the one circularity would produce---a system that drifts early enlarges its own dynamic region---and we flag it as an open validity question rather than dismissing it. Removing precipitation, the category a two-way partition cannot represent, leaves the normalized-background-flow ordering unchanged.}
\label{tab:robustness}
\end{table*}

\section{Paired Differences}
\label{sup:paired}

Separation claims in the main text are made on the bootstrap distribution of
per-prompt \emph{differences}, not on whether two marginal intervals overlap.
Every system sees the same prompts, so the paired difference is the statistic
the design supports: overlapping marginal intervals can conceal a reliable
paired difference, and non-overlapping ones can suggest a difference that is
not there.

\begin{table*}[tb]
\centering
\small
\setlength{\tabcolsep}{6pt}
\begin{tabular}{l c r c}
\toprule
Comparison & $n$ & mean diff. & 95\% CI \\
\midrule
Infinite-Forcing $-$ Causal-Forcing & 23 & -83.98 & $[-102.59,\,-65.52]$ \\
Infinite-Forcing $-$ LongLive & 23 & -10.22 & $[-14.72,\,-6.18]$ \\
Infinite-Forcing $-$ Self-Forcing & 23 & -9.07 & $[-13.29,\,-5.53]$ \\
Infinite-Forcing $-$ Rolling-Forcing & 23 & -7.58 & $[-10.38,\,-5.12]$ \\
Infinite-Forcing $-$ CausVid & 23 & -5.62 & $[-9.61,\,-2.10]$ \\
Infinite-Forcing $-$ Reward-Forcing & 23 & -4.86 & $[-7.48,\,-2.69]$ \\
\bottomrule
\end{tabular}
\caption{\textbf{Paired differences in NBF, 60s horizon.} Every system is evaluated on the same prompts, so separation is judged on the bootstrap distribution of per-prompt \emph{differences} (10k resamples, fixed seed) rather than on whether two marginal intervals happen to overlap---overlapping intervals can hide a reliable paired difference, and non-overlapping ones can suggest a difference that is not. 6 of 6 intervals exclude zero.}
\label{tab:paired}
\end{table*}

\section{Generation Configuration}
\label{sup:config}

Every text-conditioned checkpoint was loaded by the same inference program and
configuration. Fixed before the sweep to give the distilled checkpoint family
a shared four-step budget, it specifies a four-step denoising schedule with
indices $1000/750/500/250$ warped by the released scheduler at timestep shift $5.0$, six frames per
block, and seed $0$. All T2V outputs are $832\!\times\!480$ at $16$ fps. Within
a horizon, only checkpoint weights and their recorded regular/EMA selection
vary. The YAML retains a \texttt{guidance\_scale: 5.0} field, but the selected
few-step \texttt{CausalInferencePipeline} is conditional-only and never reads
it; classifier-free guidance is therefore not an applied parameter in this
sweep.

We state plainly what this does and does not license. It is a recorded
common-configuration comparison, reproducible because the inference path is
uniform and recorded. It is \emph{not} an audit of each method under its authors' released
defaults, and we no longer describe it as one. The cost is real: several of
these checkpoints are distilled for a specific step budget, and evaluating them
at a common four-step schedule does not observe them at their own operating
point. Results are therefore statements about these checkpoints under this
configuration. Re-running the sweep at each method's released defaults, and
reporting how much the ordering moves, is the natural next version of this
audit.

\begin{table*}[tb]
\centering
\small
\setlength{\tabcolsep}{5pt}
\begin{tabular}{l l c c l}
\toprule
System & setting & resolution & fps & horizons \\
\midrule
CausVid & common & 832$\times$480 & 16 & 5s, 60s, 120s, 240s \\
Self-Forcing & common & 832$\times$480 & 16 & 5s, 60s, 120s, 240s \\
Infinite-Forcing & common & 832$\times$480 & 16 & 5s, 60s, 120s, 240s \\
Rolling-Forcing & common & 832$\times$480 & 16 & 5s, 60s, 120s, 240s \\
Reward-Forcing & common & 832$\times$480 & 16 & 5s, 60s, 120s, 240s \\
LongLive & common & 832$\times$480 & 16 & 5s, 60s, 120s, 240s \\
Causal-Forcing & common & 832$\times$480 & 16 & 5s, 60s, 120s, 240s \\
\bottomrule
\end{tabular}
\caption{\textbf{Recorded generation configuration (T2V).} Every audited system was run from its released checkpoint. Output formats are normalized at measurement time: NBF is expressed per second and per frame width, fBD as a percentage of the frame diagonal, and every sequence is resampled to a common flow-estimation rate. All rows use the recorded common T2V configuration: a four-step denoising schedule with released-scheduler warping, six frames per block, and seed $0$. These are checkpoint results under this common configuration, not reconstructions of the methods' released inference procedures.}
\label{tab:native_config}
\end{table*}

The image-conditioned roster contains two distinct settings. Unmarked rows are
recorded outputs produced by model-specific released pipelines. Rows
marked $^{\dagger}$ use the common long-horizon rollout wrapper with six frames
per block and a fixed seed; unlike T2V, denoising schedules are
checkpoint-specific. These wrapper parameters
do not apply to unmarked rows. The marked rows are read as deployment
sensitivity and are never ranked against the recorded released-pipeline outputs.

\begin{table*}[tb]
\centering
\small
\setlength{\tabcolsep}{5pt}
\begin{tabular}{l l c c l}
\toprule
System & setting & resolution & fps & horizons \\
\midrule
Causal-Forcing++ (2-step, frame-wise) & released & 832$\times$480 & 16 & 5s, 60s, 120s, 240s \\
Wan2.1-I2V-14B-480P & released & 832$\times$464 & 16 & 5s \\
Wan2.2-I2V-A14B & released & 832$\times$464 & 16 & 5s \\
LTX-Video 13B-0.9.8-distilled & released & 832$\times$480 & 24 & 5s \\
Causal-Forcing++ (1-step, frame-wise) & released & 832$\times$480 & 16 & 60s, 120s \\
Causal-Forcing++ (2-step)$^{\dagger}$ & wrapper & 832$\times$480 & 16 & 5s, 60s, 120s, 240s \\
Causal-Forcing++ (1-step)$^{\dagger}$ & wrapper & 832$\times$480 & 16 & 5s, 60s, 120s, 240s \\
Causal-Forcing (frame-wise)$^{\dagger}$ & wrapper & 832$\times$480 & 16 & 5s, 60s, 120s, 240s \\
Self-Forcing$^{\dagger}$ & wrapper & 832$\times$480 & 16 & 5s, 60s, 120s, 240s \\
CausVid$^{\dagger}$ & wrapper & 832$\times$480 & 16 & 5s, 60s, 120s, 240s \\
\bottomrule
\end{tabular}
\caption{\textbf{Recorded generation configuration (I2V).} Every audited system was run from its released checkpoint. Output formats are normalized at measurement time: NBF is expressed per second and per frame width, fBD as a percentage of the frame diagonal, and every sequence is resampled to a common flow-estimation rate. Unmarked rows are recorded outputs from model-specific released pipelines and carry the setting label \emph{released}. $^{\dagger}$ rows use the common long-horizon rollout wrapper (six frames per block, fixed seed; checkpoint-specific denoising schedules); those parameters do not apply to unmarked rows, and the two settings are not ranked together.}
\label{tab:native_config_i2v}
\end{table*}

\section{Measurement Coverage}
\label{sup:coverage}

Coverage is counted from valid per-video measurements rather than from the
presence of a results file, so a run that terminated early is reported as
partial. Separately, fBD is feature-based~\cite{rublee2011orb} and declines to
report where too few repeatable keypoints survive; those abstentions are
tabulated rather than folded silently into a smaller $n$.

\begin{table}[tb]
\centering
\small
\begin{tabular}{llccc}
\toprule
track & horizon & measured & fBD abstained & rate \\
\midrule
I2V & 120s & 140 & 0 & 0.0\% \\
I2V & 240s & 30 & 0 & 0.0\% \\
I2V & 5s & 90 & 0 & 0.0\% \\
I2V & 60s & 210 & 2 & 1.0\% \\
T2V & 120s & 42 & 0 & 0.0\% \\
T2V & 240s & 28 & 0 & 0.0\% \\
T2V & 5s & 48 & 0 & 0.0\% \\
T2V & 60s & 162 & 0 & 0.0\% \\
\bottomrule
\end{tabular}
\caption{\textbf{Public-roster measurement coverage and fBD abstention.} fBD is feature-based and declines to report when too few repeatable keypoints survive in the static region. Public-roster abstentions are confined to a desert dust-storm scene whose flat, dust-obscured ground offers little stable structure to track. The remaining factors are reported for those clips as usual; only fBD is withheld.}
\label{tab:abstention}
\end{table}
\begin{table*}[!htbp]
\centering
\small
\setlength{\tabcolsep}{5pt}
\begin{tabular}{lcccccc}
\toprule
Method & status & setting & 5s & 60s & 120s & 240s \\
\midrule
CausVid & public & common & V6 T6 B & V23 T23 B & V6 T6 B & V4 T4 B \\
Self-Forcing & public & common & V6 T6 B & V23 T23 B & V6 T6 B & V4 T4 B \\
Infinite-Forcing & public & common & V6 T6 B & V23 T23 B & V6 T6 B & V4 T4 B \\
Rolling-Forcing & public & common & V6 T6 B & V24 T24 B & V6 T6 B & V4 T4 B \\
Reward-Forcing & public & common & V12 T12 B & V23 T23 B & V6 T6 B & V4 T4 B \\
LongLive & public & common & V6 T6 B & V23 T23 B & V6 T6 B & V4 T4 B \\
Causal-Forcing & public & common & V6 T6 B & V23 T23 B & V6 T6 B & V4 T4 B \\
\bottomrule
\end{tabular}
\caption{Asset and score coverage for the audited public systems. \texttt{V$n$} is the number of available video assets, \texttt{T$n$} the number of valid SNF task-metric records, and \texttt{B} denotes available VBench scores. \texttt{T$n$/$N$} marks entries where only $n$ of $N$ videos hold usable measurements. V counts generated assets, whereas audit $n$ counts unique frozen-manifest prompt items after duplicate prompt outputs are collapsed. Reward-Forcing has 12 assets over six 5\,s prompts; Rolling-Forcing has 24 assets over 23 60\,s prompts.}
\label{tab:coverage}
\end{table*}

\section{Complete Audit, Both Tracks}
\label{sup:allhorizons}

The main paper reports the 60s and 120s deployment panels; the 5s tier is an initialization check and the 240s tier is an exploratory stress horizon. Both T2V and I2V tracks appear here at every horizon, with uncertainty from a percentile bootstrap over
items~\cite{efron1979bootstrap}. For the image-conditioned (I2V) track, dagger-marked
\emph{wrapper} rows use a common long-horizon rollout wrapper rather than a
released pipeline, so their scores are read as deployment sensitivity
rather than as the published method's performance.

\begin{table*}[!tb]
\centering
\small
\setlength{\tabcolsep}{4pt}
\begin{tabular}{lccccccc}
\toprule
Method & fBD$\downarrow$ & NBF$\downarrow$ & MCFF-E & MCFF-L$\uparrow$ & FP$\uparrow$ & DLR$\downarrow$ & DAR$\downarrow$ \\
\midrule
\multicolumn{8}{l}{\textit{5s horizon  ($n=6$)}} \\
CausVid & 2.08 & 6.56 & 1.44 & 2.91 & 1.24 & 0.26 & 0.07 \\
Self-Forcing & 0.68 & 10.8 & 7.70 & 4.35 & 0.76 & 0.49 & 0.03 \\
Infinite-Forcing & 0.24 & 2.32 & 1.23 & 0.89 & 1.08 & 0.21 & 0.09 \\
Rolling-Forcing & 3.51 & 12.1 & 3.10 & 3.69 & 1.32 & 0.48 & 0.17 \\
Reward-Forcing & 1.10 & 5.21 & 0.63 & 0.58 & 1.37 & 0.44 & 0.28 \\
LongLive & 1.91 & 7.31 & 1.25 & 2.27 & 1.16 & 0.59 & 0.32 \\
Causal-Forcing & 9.55 & 31.0 & 5.32 & 3.01 & 1.13 & 0.62 & 0.37 \\
\midrule
\multicolumn{8}{l}{\textit{60s horizon  ($n=23$)}} \\
CausVid & 8.08 & 9.27 & 2.87 & 1.96 & 0.67 & 0.45 & 0.14 \\
Self-Forcing & 11.4 & 12.7 & 10.4 & 2.54 & 0.44 & 0.50 & 0.14 \\
Infinite-Forcing & 7.75 & 3.65 & 1.17 & 0.29 & 0.63 & 0.55 & 0.14 \\
Rolling-Forcing & 16.8 & 11.2 & 3.18 & 1.57 & 0.75 & 0.54 & 0.18 \\
Reward-Forcing & 14.8 & 8.51 & 3.81 & 0.80 & 0.49 & 0.60 & 0.17 \\
LongLive & 15.9 & 13.9 & 6.46 & 2.41 & 0.75 & 0.59 & 0.17 \\
Causal-Forcing & 22.4 & 87.6 & 9.84 & 7.21 & 0.83 & 0.94 & 0.43 \\
\midrule
\multicolumn{8}{l}{\textit{120s horizon  ($n=6$)}} \\
CausVid & 6.15 & 3.42 & 0.36 & 0.52 & 1.06 & 0.41 & 0.13 \\
Self-Forcing & 21.2 & 6.80 & 4.85 & 0.82 & 0.85 & 0.75 & 0.15 \\
Infinite-Forcing & 16.1 & 2.60 & 0.50 & 0.32 & 0.64 & 0.57 & 0.21 \\
Rolling-Forcing & 23.8 & 10.6 & 3.33 & 0.77 & 0.38 & 0.82 & 0.22 \\
Reward-Forcing & 20.0 & 11.6 & 2.05 & 0.69 & 0.39 & 0.70 & 0.12 \\
LongLive & 22.5 & 9.66 & 5.10 & 1.19 & 0.81 & 0.84 & 0.23 \\
Causal-Forcing & 21.8 & 130 & 10.5 & 10.7 & 0.90 & 0.81 & 0.50 \\
\midrule
\multicolumn{8}{l}{\textit{240s horizon  ($n=4$)}} \\
CausVid & 7.23 & 13.4 & 0.42 & 0.87 & 0.99 & 0.85 & 0.08 \\
Self-Forcing & 20.0 & 17.4 & 3.09 & 0.50 & 0.15 & 2.26 & 0.20 \\
Infinite-Forcing & 20.0 & 4.15 & 0.33 & 0.23 & 1.02 & 1.08 & 0.15 \\
Rolling-Forcing & 23.7 & 15.4 & 2.15 & 0.97 & 0.46 & 0.78 & 0.18 \\
Reward-Forcing & 26.4 & 11.4 & 1.09 & 3.55 & 1.53 & 0.50 & 0.07 \\
LongLive & 23.1 & 11.8 & 1.52 & 1.32 & 1.20 & 1.38 & 0.31 \\
Causal-Forcing & 18.9 & 102 & 4.72 & 6.90 & 1.21 & 0.73 & 0.58 \\
\bottomrule
\end{tabular}
\caption{\textbf{Complete T2V audit, all four horizons.} The main paper reports the 60\,s and 120\,s panels; the 5\,s tier is an initialization check and 240\,s a diagnostic extreme. Every row is computed under the same frozen specification, so panels are comparable within a horizon; prompt sets differ across horizons, so columns are not comparable between panels. Header $n$ is the maximum valid item count within each horizon; factor-specific abstentions are in Table~S6. Units: fBD is \% frame diagonal; NBF is $10^{-3}$ frame widths/s; MCFF is px/sampled interval; FP is unitless. FP is the mean of per-video clipped late/early ratios and therefore need not equal the ratio of the displayed aggregate MCFF-L and MCFF-E means.}
\label{tab:t2v_audit_full}
\end{table*}
\begin{table*}[!tb]
\centering
\small
\setlength{\tabcolsep}{4pt}
\begin{tabular}{llccccccc}
\toprule
Method & setting & fBD$\downarrow$ & NBF$\downarrow$ & MCFF-E & MCFF-L$\uparrow$ & FP$\uparrow$ & DLR$\downarrow$ & DAR$\downarrow$ \\
\midrule
\multicolumn{9}{l}{\textit{5s horizon  ($n=10$)}} \\
Causal-Forcing++ (2-step) & wrapper & 3.69 & 10.0 & 1.06 & 0.61 & 0.78 & 0.67 & 0.32 \\
Causal-Forcing++ (1-step) & wrapper & 5.28 & 11.4 & 1.05 & 0.47 & 0.64 & 0.66 & 0.29 \\
Causal-Forcing (frame-wise) & wrapper & 16.1 & 45.3 & 8.48 & 2.79 & 0.43 & 0.68 & 0.29 \\
Causal-Forcing++ (2-step, frame-wise) & released & 1.22 & 10.6 & 2.23 & 0.78 & 0.47 & 0.68 & 0.20 \\
Self-Forcing & wrapper & 9.66 & 5.74 & 0.69 & 0.26 & 0.70 & 0.74 & 0.21 \\
CausVid & wrapper & 15.2 & 10.8 & 0.76 & 0.87 & 0.98 & 1.04 & 0.10 \\
Wan2.1-I2V-14B-480P & released & 5.01 & 21.3 & 6.67 & 5.58 & 1.09 & 0.52 & 0.18 \\
Wan2.2-I2V-A14B & released & 6.31 & 25.3 & 4.33 & 9.09 & 0.95 & 0.50 & 0.27 \\
LTX-Video 13B-0.9.8-distilled & released & 6.65 & 17.8 & 3.59 & 3.48 & 0.92 & 0.39 & 0.24 \\
\midrule
\multicolumn{9}{l}{\textit{60s horizon  ($n=30$)}} \\
Causal-Forcing++ (2-step) & wrapper & 9.18 & 5.97 & 0.85 & 0.60 & 0.86 & 0.67 & 0.25 \\
Causal-Forcing++ (1-step) & wrapper & 5.34 & 7.05 & 0.86 & 0.57 & 0.77 & 0.64 & 0.28 \\
Causal-Forcing (frame-wise) & wrapper & 21.1 & 36.4 & 2.80 & 2.69 & 0.89 & 0.80 & 0.35 \\
Causal-Forcing++ (2-step, frame-wise) & released & 9.94 & 6.91 & 1.30 & 0.78 & 1.02 & 0.75 & 0.17 \\
Self-Forcing & wrapper & 17.3 & 5.52 & 1.50 & 0.69 & 0.89 & 0.67 & 0.18 \\
CausVid & wrapper & 19.9 & 3.31 & 0.58 & 0.41 & 0.81 & 0.57 & 0.14 \\
Causal-Forcing++ (1-step, frame-wise) & released & 10.4 & 5.25 & 0.53 & 0.52 & 1.01 & 0.83 & 0.22 \\
\midrule
\multicolumn{9}{l}{\textit{120s horizon  ($n=20$)}} \\
Causal-Forcing++ (2-step) & wrapper & 9.65 & 8.81 & 0.89 & 0.82 & 0.93 & 0.58 & 0.28 \\
Causal-Forcing++ (1-step) & wrapper & 7.26 & 9.39 & 0.84 & 0.66 & 0.86 & 0.62 & 0.31 \\
Causal-Forcing (frame-wise) & wrapper & 19.9 & 52.7 & 4.82 & 5.06 & 1.08 & 0.67 & 0.24 \\
Causal-Forcing++ (2-step, frame-wise) & released & 11.4 & 17.3 & 1.89 & 1.54 & 1.01 & 0.75 & 0.09 \\
Self-Forcing & wrapper & 23.0 & 10.3 & 2.41 & 0.76 & 0.45 & 0.66 & 0.26 \\
CausVid & wrapper & 19.7 & 6.04 & 1.15 & 0.83 & 0.67 & 0.48 & 0.14 \\
Causal-Forcing++ (1-step, frame-wise) & released & 13.6 & 7.39 & 1.01 & 0.35 & 0.64 & 0.77 & 0.33 \\
\midrule
\multicolumn{9}{l}{\textit{240s horizon  ($n=5$)}} \\
Causal-Forcing++ (2-step) & wrapper & 10.4 & 4.83 & 0.53 & 0.59 & 1.05 & 0.55 & 0.15 \\
Causal-Forcing++ (1-step) & wrapper & 6.42 & 6.66 & 0.56 & 0.51 & 0.83 & 0.67 & 0.30 \\
Causal-Forcing (frame-wise) & wrapper & 19.8 & 31.5 & 3.08 & 2.28 & 0.74 & 0.69 & 0.45 \\
Causal-Forcing++ (2-step, frame-wise) & released & 6.32 & 3.95 & 1.23 & 0.68 & 0.67 & 0.83 & 0.02 \\
Self-Forcing & wrapper & 22.3 & 5.74 & 1.48 & 1.59 & 0.81 & 0.71 & 0.08 \\
CausVid & wrapper & 18.4 & 2.08 & 0.81 & 0.79 & 0.86 & 0.44 & 0.11 \\
\bottomrule
\end{tabular}
\caption{\textbf{Complete I2V audit, all four horizons.} The main paper reports the 60s and 120s panels; the 5s tier is an initialization check and the 240s tier is an exploratory stress horizon. All rows are scored with the same frozen measurement specification; however, released-pipeline and wrapper rows represent different generation settings and are reported separately rather than ranked against one another. Prompt sets differ across horizons, so values are not compared across panels. Header $n$ is the maximum valid item count within each horizon; factor-specific abstentions are in Table~S6. Units: fBD is \% frame diagonal; NBF is $10^{-3}$ frame widths/s; MCFF is px/sampled interval; FP is unitless. FP is the mean of per-video clipped late/early ratios and therefore need not equal the ratio of the displayed aggregate MCFF-L and MCFF-E means.}
\label{tab:i2v_audit_full}
\end{table*}

\begin{figure}[tb]
\centering
\includegraphics[width=\linewidth]{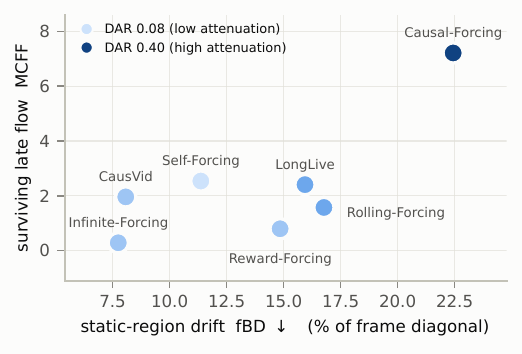}
\caption{\textbf{Operating regimes under the recorded common T2V configuration.} Each checkpoint in the static-drift--surviving-flow plane. Color encodes low-to-high attenuation of measured dynamic-region flow after global-motion compensation, not a share attributable to drift: the quantity is signed, and values are shown clipped to $[0,1]$. The view exposes trade-offs; it is not a leaderboard.}
\label{sup:fig_regimes}
\end{figure}

\section{Aggregation Sensitivity}
\label{sup:aggregation}

The evaluation set is dominated by one flow medium, so the audit must be shown
not to be a report on that medium alone. The main paper aggregates by averaging
over prompts. The alternative, weighting each scene category equally, is not
preferable here: at the principal horizon three of the five populated categories
contain two prompts each, so a macro-average would give a two-prompt cell the
same influence as a fourteen-prompt one and inflate variance in exchange for
nominal balance. We therefore report both and state what changes.

The best- and worst-ranked systems are the same under either aggregation, so the
contrast the paper draws does not depend on the choice. Intermediate ranks do
move, which is why no ordinal claim is made about the middle of the field.

\begin{table*}[tb]
\centering
\small
\setlength{\tabcolsep}{5pt}
\begin{tabular}{l c c c c}
\toprule
System & prompt mean & rank & macro mean & rank \\
\midrule
\multicolumn{5}{l}{\emph{NBF}} \\
\quad Infinite-Forcing & 3.65 & 1 & 4.12 & 1 \\
\quad Reward-Forcing & 8.51 & 2 & 10.11 & 4$^{\dagger}$ \\
\quad CausVid & 9.27 & 3 & 9.93 & 3 \\
\quad Rolling-Forcing & 11.23 & 4 & 11.32 & 5$^{\dagger}$ \\
\quad Self-Forcing & 12.72 & 5 & 8.98 & 2$^{\dagger}$ \\
\quad LongLive & 13.86 & 6 & 14.69 & 6 \\
\quad Causal-Forcing & 87.62 & 7 & 61.70 & 7 \\
\multicolumn{5}{l}{\emph{fBD}} \\
\quad Infinite-Forcing & 7.75 & 1 & 6.10 & 1 \\
\quad CausVid & 8.08 & 2 & 9.51 & 3$^{\dagger}$ \\
\quad Self-Forcing & 11.37 & 3 & 9.27 & 2$^{\dagger}$ \\
\quad Reward-Forcing & 14.85 & 4 & 15.61 & 5$^{\dagger}$ \\
\quad LongLive & 15.94 & 5 & 14.93 & 4$^{\dagger}$ \\
\quad Rolling-Forcing & 16.77 & 6 & 18.85 & 6 \\
\quad Causal-Forcing & 22.44 & 7 & 23.49 & 7 \\
\bottomrule
\end{tabular}
\caption{\textbf{Sensitivity of the audit to category aggregation.} Prompt means, as reported in the main paper, against category macro-averages, for the two static-fidelity factors at the principal horizon. $^{\dagger}$ marks a system whose rank changes. The best- and worst-ranked systems are identical under both, so the paper's headline contrast does not depend on the choice; intermediate positions do change, which is why no ordinal claim is made about the middle of the field.}
\label{tab:aggregation}
\end{table*}

\section{Whole-Frame and SNF-Bench Rank Correlation}
\label{sup:disagreement}
Tables~\ref{tab:disagreement} and~\ref{tab:interpretation_changes} report the method-level rank correlations and all interpretation changes satisfying the fixed inclusion rule.

\begin{table*}[htb]
\centering
\small
\setlength{\tabcolsep}{3pt}
\begin{tabular}{lcccc}
\toprule
generic metric & fBD & NBF & MCFF & DLR \\
\midrule
VB-DD & +0.75 & +0.93 & +0.89 & +0.36 \\
VB-bg & -0.64 & -0.71 & -0.75 & -0.50 \\
VB-smooth / VB-flick & -0.96 & -0.82 & -0.61 & -0.64 \\
\bottomrule
\end{tabular}
\caption{\textbf{Rank disagreement at 60\,s.} Spearman correlation over $n=7$ method means---descriptive, not inferential. Positive correlation between apparent motion and background flow is the effect the benchmark isolates. Smoothness and flickering rank these systems identically and share a row.}
\label{tab:disagreement}
\end{table*}
\begin{table*}[t]
\centering
\small
\begin{tabular}{l l l p{0.35\textwidth}}
\toprule
Method & Generic evidence & SNF-Bench evidence & Interpretation \\
\midrule
CausVid & VB-DD rank 6/7 & NBF rank 3/7, MCFF-L rank 4/7 & stable sequence with decaying intended motion \\
Causal-Forcing & VB-DD rank 1/7 & NBF rank 7/7, MCFF-L rank 1/7 & high apparent motion accompanied by high static-region motion under the recorded common configuration \\
Infinite-Forcing & VB-DD rank 7/7 & NBF rank 1/7, MCFF-L rank 7/7 & stable sequence with decaying intended motion \\
LongLive & VB-DD rank 3/7 & NBF rank 6/7, MCFF-L rank 3/7 & high apparent motion accompanied by high static-region motion under the recorded common configuration \\
Reward-Forcing & VB-DD rank 5/7 & NBF rank 2/7, MCFF-L rank 6/7 & stable sequence with decaying intended motion \\
Self-Forcing & VB-DD rank 2/7 & NBF rank 5/7, MCFF-L rank 2/7 & high apparent motion accompanied by high static-region motion under the recorded common configuration \\
\bottomrule
\end{tabular}
\caption{\textbf{Interpretation changes between generic whole-frame metrics and SNF-Bench at 60\,s.} Neither evaluation is labeled correct; the table identifies information hidden by whole-frame aggregation. Cases are selected by a fixed rank-gap rule applied to all public methods. Rolling-Forcing does not satisfy the fixed three-rank-gap inclusion criterion and is therefore absent.}
\label{tab:interpretation_changes}
\end{table*}

\section{Deployment Sensitivity}
\label{sup:deployment}
Table~\ref{tab:deployment_sensitivity} reports changes from released-pipeline outputs to the common long-horizon wrapper for checkpoints evaluated under both settings.

\begin{table*}[tb]
\centering
\small
\setlength{\tabcolsep}{6pt}
\begin{tabular}{l c c c c c}
\toprule
Checkpoint & horizon & $\Delta$fBD & $\Delta$NBF & $\Delta$MCFF-L & $\Delta$DLR \\
\midrule
Causal-Forcing++ (2-step) & 5s & +2.47 & -0.61 & -0.17 & -0.01 \\
Causal-Forcing++ (2-step) & 60s & -0.76 & -0.95 & -0.17 & -0.08 \\
Causal-Forcing++ (2-step) & 120s & -1.71 & -8.46 & -0.73 & -0.17 \\
Causal-Forcing++ (2-step) & 240s & +4.06 & +0.88 & -0.09 & -0.28 \\
Causal-Forcing++ (1-step) & 60s & -5.05 & +1.80 & +0.05 & -0.20 \\
Causal-Forcing++ (1-step) & 120s & -6.34 & +2.00 & +0.31 & -0.14 \\
\bottomrule
\end{tabular}
\caption{\textbf{Deployment sensitivity.} Change from each checkpoint's own released setting to the common long-horizon wrapper, for the two checkpoints evaluated both ways. Absolute wrapper scores are never read as the published method's performance.}
\label{tab:deployment_sensitivity}
\end{table*}

\section{Signed Drift Attenuation}
\label{sup:dar}
Table~\ref{tab:darneg} reports the incidence and magnitude of negative signed DAR values.

\begin{table*}[!htbp]
\centering
\small
\setlength{\tabcolsep}{5pt}
\begin{tabular}{lcccccc}
\toprule
Method & track & dur & n & negative & rate & most negative \\
\midrule
CausVid & i2v & 120s & 20 & 1 & 5.0\% & -0.055 \\
CausVid & i2v & 5s & 10 & 5 & 50.0\% & -1.076 \\
CausVid & i2v & 60s & 30 & 2 & 6.7\% & -0.084 \\
Causal-Forcing++ (1-step) & i2v & 120s & 20 & 1 & 5.0\% & -0.027 \\
Causal-Forcing (frame-wise) & i2v & 120s & 20 & 3 & 15.0\% & -0.171 \\
Causal-Forcing++ (2-step) & i2v & 5s & 10 & 1 & 10.0\% & -0.164 \\
Causal-Forcing++ (1-step, frame-wise) & i2v & 60s & 30 & 3 & 10.0\% & -1.420 \\
Causal-Forcing++ (2-step, frame-wise) & i2v & 120s & 20 & 5 & 25.0\% & -0.241 \\
Causal-Forcing++ (2-step, frame-wise) & i2v & 240s & 5 & 1 & 20.0\% & -0.651 \\
Causal-Forcing++ (2-step, frame-wise) & i2v & 5s & 10 & 1 & 10.0\% & -0.288 \\
Causal-Forcing++ (2-step, frame-wise) & i2v & 60s & 30 & 1 & 3.3\% & -0.001 \\
LTX-Video 13B-0.9.8-distilled & i2v & 5s & 10 & 2 & 20.0\% & -0.018 \\
Self-Forcing & i2v & 120s & 20 & 2 & 10.0\% & -0.046 \\
Self-Forcing & i2v & 240s & 5 & 1 & 20.0\% & -0.136 \\
Self-Forcing & i2v & 60s & 30 & 4 & 13.3\% & -0.578 \\
Wan2.1-I2V-14B-480P & i2v & 5s & 10 & 1 & 10.0\% & -0.255 \\
Wan2.2-I2V-A14B & i2v & 5s & 10 & 1 & 10.0\% & -0.335 \\
CausVid & t2v & 240s & 4 & 1 & 25.0\% & -0.384 \\
CausVid & t2v & 5s & 6 & 2 & 33.3\% & -0.093 \\
Causal-Forcing & t2v & 5s & 6 & 2 & 33.3\% & -0.273 \\
Causal-Forcing & t2v & 60s & 23 & 3 & 13.0\% & -0.640 \\
Infinite-Forcing & t2v & 240s & 4 & 1 & 25.0\% & -0.158 \\
Infinite-Forcing & t2v & 60s & 23 & 4 & 17.4\% & -0.359 \\
LongLive & t2v & 120s & 6 & 1 & 16.7\% & -1.076 \\
LongLive & t2v & 5s & 6 & 1 & 16.7\% & -0.139 \\
LongLive & t2v & 60s & 23 & 2 & 8.7\% & -0.051 \\
Reward-Forcing & t2v & 120s & 6 & 1 & 16.7\% & -0.116 \\
Reward-Forcing & t2v & 5s & 12 & 2 & 16.7\% & -0.145 \\
Reward-Forcing & t2v & 60s & 23 & 3 & 13.0\% & -0.317 \\
Rolling-Forcing & t2v & 120s & 6 & 3 & 50.0\% & -0.353 \\
Rolling-Forcing & t2v & 5s & 6 & 1 & 16.7\% & -0.068 \\
Rolling-Forcing & t2v & 60s & 24 & 2 & 8.3\% & -0.038 \\
Self-Forcing & t2v & 120s & 6 & 2 & 33.3\% & -0.130 \\
Self-Forcing & t2v & 240s & 4 & 1 & 25.0\% & -0.153 \\
Self-Forcing & t2v & 5s & 6 & 1 & 16.7\% & -0.046 \\
Self-Forcing & t2v & 60s & 23 & 4 & 17.4\% & -1.117 \\
\bottomrule
\end{tabular}
\caption{Incidence of negative DAR, i.e.\ clips where global-motion compensation \emph{increased} measured dynamic-region flow. 21 further entries had no negative values and are omitted.}
\label{tab:darneg}
\end{table*}

Drift attenuation is stored signed and reported clipped to $[0,1]$. Negative
values are informative rather than anomalous: where local flow opposes the
estimated global field, compensation increases the measured magnitude. We report
their incidence rather than suppressing it.

\section{Evaluation-Set Composition}
\label{sup:categories}
Table~\ref{tab:category} summarizes category coverage by track and horizon.

\begin{table*}[!htbp]
\centering
\scriptsize
\setlength{\tabcolsep}{5pt}
\begin{tabular}{lcccccccc}
\toprule
track/duration & n & river\_stream & ocean\_waves & precipitation & fire\_smoke & lava\_volcanic & windborne & cats \\
\midrule
i2v/5s & 10 & 2 & · & 3 & 2 & 1 & 2 & 5 \\
i2v/60s & 30 & 9 & 4 & 9 & 1 & 3 & 4 & 6 \\
i2v/120s & 20 & 6 & 3 & 3 & 3 & 1 & 4 & 6 \\
i2v/240s & 5 & 1 & 2 & 1 & · & 1 & · & 4 \\
t2v/5s & 6 & 3 & 1 & 1 & 1 & · & · & 4 \\
t2v/60s & 23 & 14 & 2 & 3 & 2 & · & 2 & 5 \\
t2v/120s & 6 & 4 & · & 1 & · & · & 1 & 3 \\
t2v/240s & 4 & 2 & · & 1 & · & · & 1 & 3 \\
\bottomrule
\end{tabular}
\caption{Scene-category balance of the evaluation set, by track and horizon. A dot marks a category absent from that cell.}
\label{tab:category}
\end{table*}

\section{Additional Qualitative Material}
\label{sup:qualitative}

\paragraph{Figure selection rule.}
For each prompt represented by at least six systems, the support-preserving
candidate minimizes $\mathrm{fBD}-2\,\mathrm{MCFF\mbox{-}L}$, the drift
candidate maximizes fBD, and the stagnation candidate minimizes MCFF-L. Among
prompts with three distinct candidates, the selected prompt maximizes
$(\mathrm{fBD}_{\mathrm{drift}}-\mathrm{fBD}_{\mathrm{support}})/
(1+40|\mathrm{DD}^{*}_{\mathrm{drift}}-\mathrm{DD}^{*}_{\mathrm{support}}|)$.

\begin{figure*}[htb]
\centering
\includegraphics[width=\textwidth]{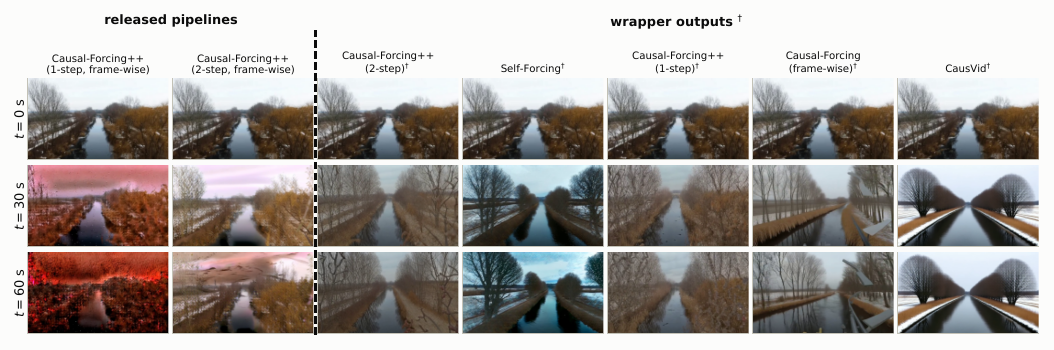}
\caption{\textbf{Image-conditioned systems on one shared source image.} Every
system begins from the identical frame, so divergence over the rollout is
attributable to the system rather than to a differently imagined scene---which
is not true of the text-conditioned track, where each system authors its own
layout. Released-pipeline and wrapper outputs are visibly separated and ordered
by increasing fBD only within their respective groups; $^{\dagger}$ marks a
wrapper output, and no cross-setting ordering is intended.}
\label{sup:fig_i2v}
\end{figure*}

\begin{figure*}[htb]
\centering
\includegraphics[width=\textwidth]{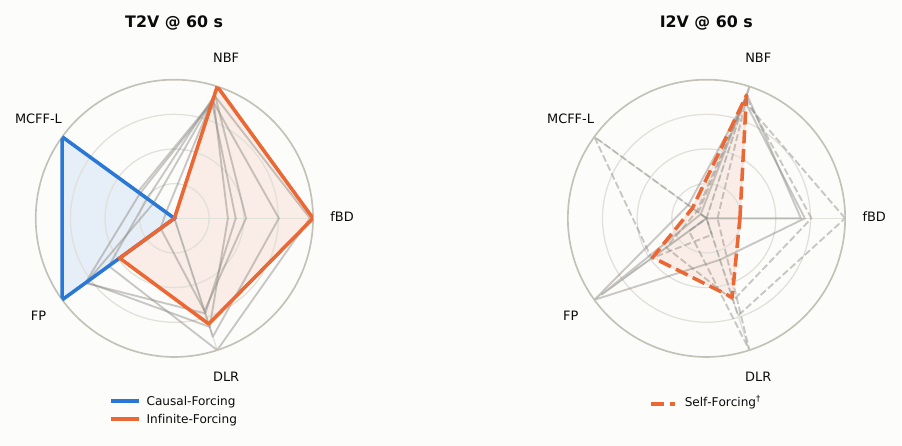}
\caption{\textbf{Factor profile per track.} Each factor is rescaled across the
audited systems and oriented so that outward is always better; plotting raw
values would place ``most drift'' and ``most surviving motion'' both outward and
the shape would carry no meaning. No system encloses the others on either track,
which is the conclusion the operating-regime view reaches, shown as a profile.
On the image-conditioned panel, dashed outlines marked $^{\dagger}$ are rows run
under the common long-horizon wrapper rather than a released pipeline: they
are shown for shape, and are not peers of the solid released-pipeline outlines. The
rescaling is relative to the audited roster, so a profile is descriptive of this
set and would change if the set changed.}
\label{sup:fig_radar}
\end{figure*}

\begin{figure*}[htb]
\centering
\includegraphics[width=\textwidth]{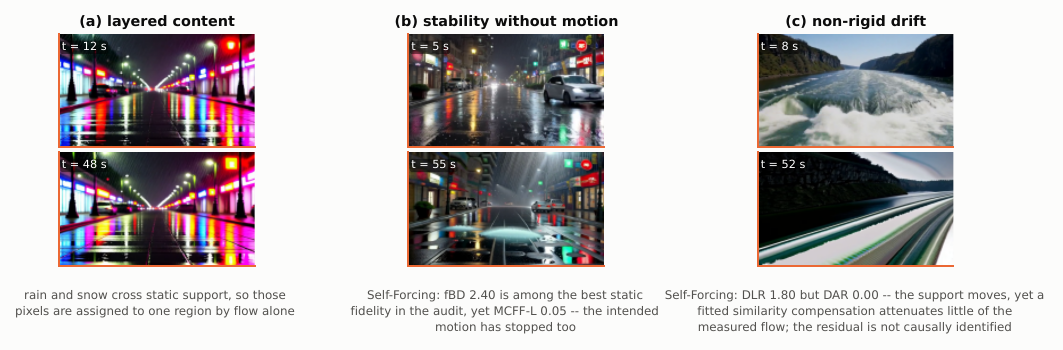}
\caption{\textbf{Stated scope boundaries.} (a)~A two-way partition cannot
separate precipitation from the support it crosses. (b)~A system can attain
near-best static fidelity precisely because its intended motion has also
stopped. (c)~High leakage with low attenuation marks static-region motion for which the
fitted global similarity model attenuates little measured flow; what the residual consists of is not identified by these measurements.}
\label{sup:fig_limits}
\end{figure*}

\begin{figure*}[htb]
\centering
\includegraphics[width=\textwidth]{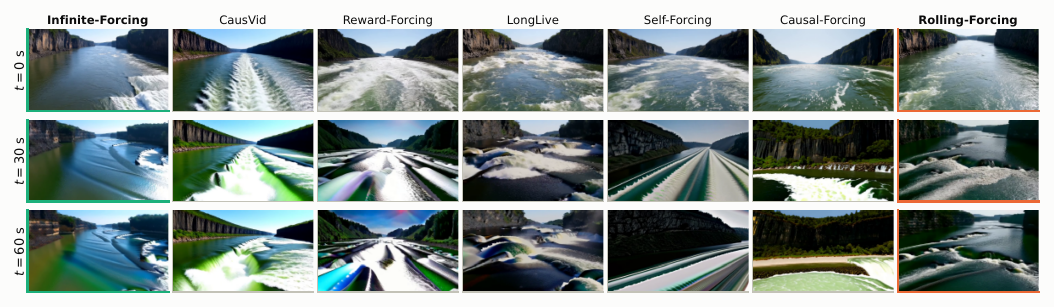}
\caption{\textbf{T2V comparison under the recorded common configuration.}
Every audited text-conditioned checkpoint is evaluated on the same prompt. Rows show t={0,30,60} s; columns are ordered by increasing measured fBD.}
\label{sup:fig_qualitative}
\end{figure*}

\begin{figure*}[tb]
\centering
\includegraphics[width=0.75\textwidth]{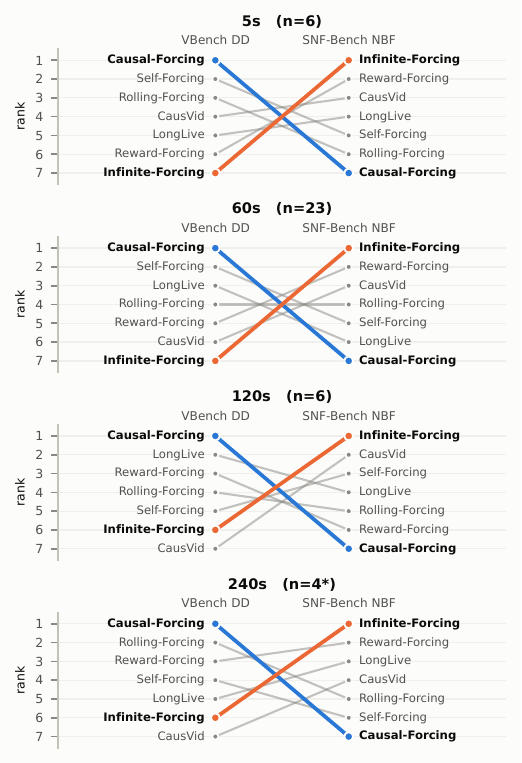}
\caption{\textbf{Rank disagreement at every audited horizon.} Rank under a
whole-frame motion score against rank under normalized background flow. Both
quantities are independent of the global-motion estimator; prompt counts are
given per panel. An asterisk marks a horizon where Dynamic Degree is available
only as a method-level VBench aggregate, suitable for descriptive ranks but not
for item-level uncertainty.}
\label{sup:fig_ranks}
\end{figure*}

\begin{figure*}[htb]
\centering
\includegraphics[width=\textwidth]{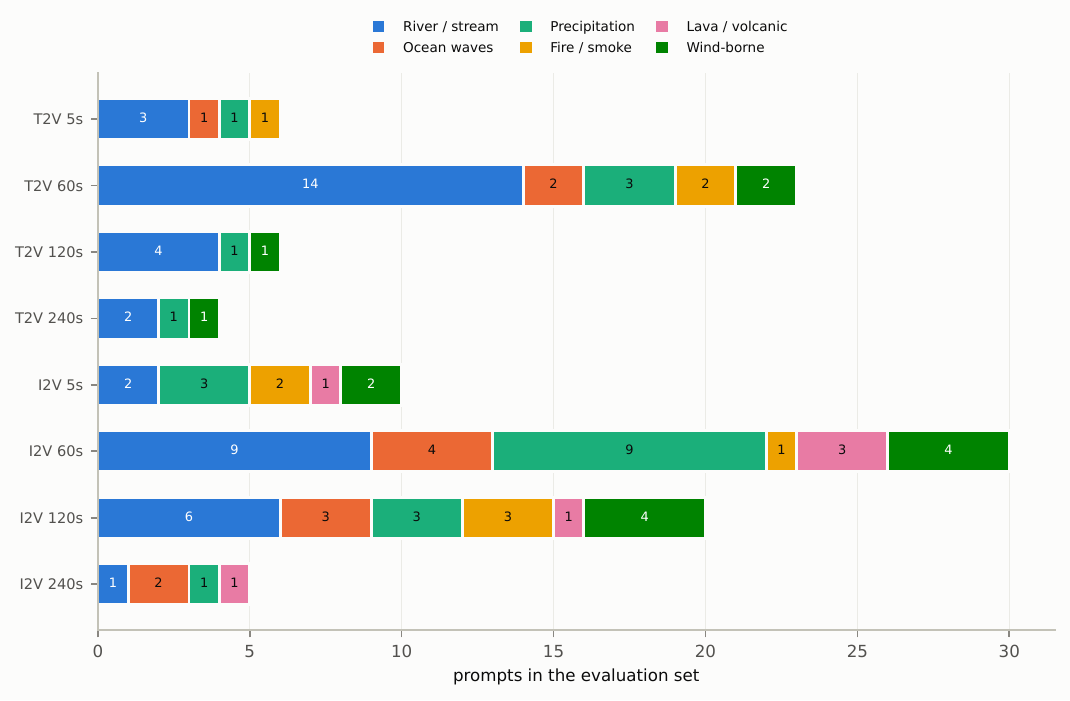}
\caption{\textbf{Composition of the evaluation set by scene category,} in
prompts per category at each track and horizon.}
\label{sup:fig_categories}
\end{figure*}

\clearpage

\end{document}